\documentclass[11pt]{article}
\usepackage[preprint]{acl}
\usepackage{times}
\usepackage{latexsym}
\usepackage[T1]{fontenc}
\usepackage[utf8]{inputenc}
\usepackage{amsmath,amssymb}
\usepackage{booktabs}
\usepackage{graphicx}
\usepackage{placeins}
\usepackage{multirow}
\newsavebox{\agritablebox}

\usepackage{xcolor}
\usepackage{natbib}
\usepackage{url}
\usepackage{xurl}
\hypersetup{hidelinks,
  pdftitle={AgriJev: Efficient Agricultural Candidate Scoring with Multi-Scale Adapted Models},
  pdfauthor={Xuechen Li},
  pdfsubject={Agricultural candidate scoring and inference evaluation},
  pdfkeywords={agricultural NLP, candidate scoring, candidate-conditioned classification, parameter-efficient fine-tuning, efficient inference, zero-shot transfer}
}
\title{AgriJev: Efficient Agricultural Candidate Scoring\\with Multi-Scale Adapted Models}
\author{Xuechen Li\\
University of Minnesota Twin Cities\\
\texttt{li003487@umn.edu}}
\date{}

\begin{document}\maketitle
\begin{abstract}
Agricultural systems make decisions over heterogeneous entity types, textual judgments, and knowledge-question options. Generating a textual answer adds decoding and extraction steps to decisions that ultimately require a candidate choice. We introduce AgriJev, a family of adapted Qwen3.5 models at 0.8B, 2B, 4B, and 9B. A shared interface conditions on agricultural text, a judgment question, and candidate definitions, returning scores and relative probabilities in one forward call. We prepare source-aware training records and frozen evaluations from existing agricultural resources, retaining original groups and annotation provenance. Comparisons include original Qwen3.5 models and matched agricultural SFT models, both evaluated through native answer generation. Across 17 external test views, AgriJev-9B reaches 75.44\% mean accuracy, exceeding the original and SFT generators by 5.89 and 3.79 percentage points. It outperforms the original generator on 13 views and narrows the mean gap to hosted Jev 1.13.0. On a fixed A100 workload, its median scoring latency is approximately half that of same-scale answer generation. Task-dependent gains and failures characterize a practical choice among agricultural model weights and inference contracts.
\end{abstract}
\begin{figure}[t]\centering
\includegraphics[width=\columnwidth]{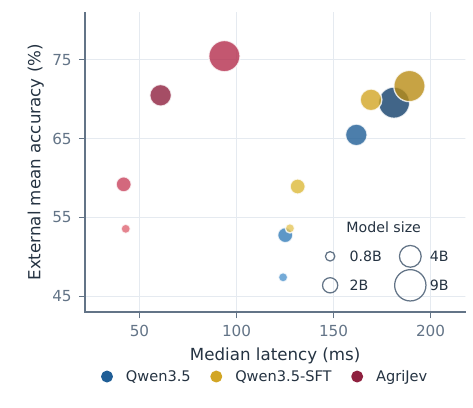}
\caption{\textbf{Accuracy, latency, and model size.} Mean accuracy uses 17 external views, excluding paired CROP reordering. p50 latency uses 32 BIO texts, 27 candidates, A100 40GB, batch 1. Qwen3.5 and Qwen3.5-SFT generate answers; AgriJev scores candidates. Bubble area is proportional to measured loaded parameters; the lower-right size key gives nominal scales. Accuracy and latency use distinct fixed workloads.}
\label{fig:external_teaser}\end{figure}

\section{Introduction}
Agricultural language processing requires decisions about both textual evidence and domain knowledge. A system may identify an annotated mention's type, check a type proposition, determine whether an event is reported, or select a knowledge-question answer. These decisions involve different candidate inventories. Entity types follow source ontologies, event judgments depend on evidence categories, and knowledge questions provide their own options. A reusable decision interface should condition explicitly on the question and the meanings of those candidates.

Agricultural resources already support parts of this setting. CROP provides crop-science resources, AgriEval assesses agricultural knowledge, and SeedBench targets seed-science questions and reading tasks \citep{zhang2024,yan2026,ying2025seedbench}. They differ in answer format, annotation provenance, and scoring unit. Judgments derived from shared extraction annotations also differ from independently authored questions. A common evaluation should preserve source groups and labels, disclose silver supervision, and retain task-specific metrics. Otherwise, aggregated results can obscure both the evaluated capability and the amount of independent evidence.

Generative models can express these decisions as short answers, but their deployment includes autoregressive decoding and answer extraction. A generated code can also fall outside the supplied candidate inventory. Selection-oriented methods provide a complementary interface for scoring described candidates directly. TARS, UniMC, flexible-label classification, and prompt-based label prediction establish relevant foundations \citep{halder2020,yang2022,stepanov2025,gaolm2021}. Symbol-binding studies further examine the connection between candidate meanings and output codes \citep{wei2023,xue2024}. The agricultural question is empirical: how do adaptation, model scale, and the inference contract affect accuracy and cost across heterogeneous source tasks?

AgriJev addresses this question with locally adapted Qwen3.5 weights and a shared text--question--candidate interface. It gathers logits for locally assigned candidate codes and returns scores and relative probabilities in one forward call. We compare this pipeline with original Qwen3.5 answer generation and agricultural SFT that supervises complete answer codes and end-of-sequence tokens. All comparisons retain the same frozen texts, questions, candidate meanings, and source answer keys. The inference prompts and output contracts are specified separately rather than treated as identical implementations.

Our contributions are threefold. First, we provide documented agricultural record preparation and an evaluation suite that preserves existing source annotations and original-context groups. Second, we instantiate a shared discriminative interface that returns candidate scores and relative probabilities without constructing an answer sequence. Third, we provide adapted weights at four scales and completed internal, external, and cost comparisons with native generated-answer references. The suite consolidates existing resources; it does not constitute a newly expert-annotated agricultural corpus.

\section{Related Work}
\paragraph{Agricultural resources and models.}
ChatAgri studies agricultural text classification, and CROP supplies crop-science adaptation and evaluation resources \citep{zhao2023,zhang2024}. AgriEval, SeedBench, and BhashaBench-Krishi cover agricultural knowledge, seed science, and multilingual questions \citep{yan2026,ying2025seedbench,devane2025}. Entity resources provide complementary annotated-mention judgments \citep{dsouza2024agriner,abdelmalak2025fairagro}. Our preparation retains their different ontologies, formats, and annotation origins. Akshara and AgriReason offer agricultural adaptation \citep{cropin2024akshara,agrireasoncard}; our candidate-scoring references do not reproduce their unrestricted advisory workflows.

\paragraph{Shared interfaces and label semantics.}
Entailment classification represents candidate descriptions as hypotheses \citep{yin2019benchmark,laurer2023universal}. TARS, UniMC, and label-description training condition decisions on label meanings \citep{halder2020,yang2022,gao2023}. GLiClass supports flexible labels, GLiNER2 combines schema-based classification and extraction, and embeddings support description matching \citep{stepanov2025,zaratiana2025,reimers2019sbert,xiao2024cpack}.

\paragraph{Prompt classification and selection.}
PET and LM-BFF adapt models through cloze tasks and label-word predictions \citep{schick2021,gaolm2021}. Symbol tuning and strengthened symbol binding study arbitrary output mappings and option-code associations \citep{wei2023,xue2024}. Jev exemplifies a selection interface over state, instructions, and described options \citep{typesafejev,almeida2026jev}. AgriJev follows this organization for agricultural decisions; its contribution concerns preparation, adapted weights, and evaluation rather than inventing label-conditioned selection. Interface similarity does not imply reproducing Jev training or establishing calibration.

\paragraph{Adaptation and evaluation contracts.}
Domain-adaptive pretraining motivates domain specialization \citep{gururangan2020}, while LoRA supports economical adaptation \citep{hu2022}. We instead supervise agricultural decisions in Qwen3.5 models \citep{qwen2026}, comparing original generation, native-answer SFT, and scoring. Task-definition and definition-receptivity studies motivate explicit questions and candidate meanings \citep{yin2023,mohammadi2025}. Option-ID preferences and reordering sensitivity require paired diagnostics \citep{zheng2024,pezeshkpour2024}; text-answer evaluation can also disagree with first-token scoring \citep{wang2024}. We therefore distinguish inference contracts, retain paired CROP views, and separate option decisions from question-level exact match.

\section{Task and Model}
\begin{figure*}[t]\centering
\includegraphics[width=\textwidth]{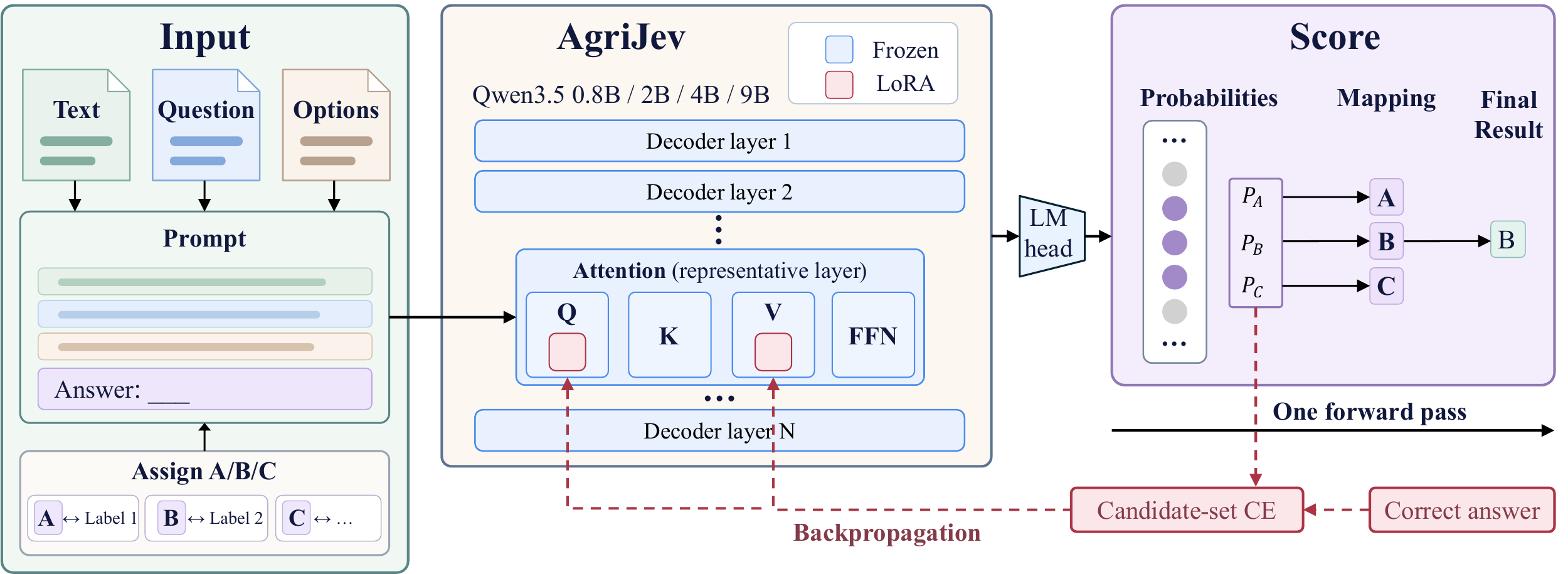}
\caption{\textbf{AgriJev candidate scoring and adaptation.} Agricultural text, a question, and candidate definitions form a prompt with locally assigned option codes. Trainable query/value LoRA augments frozen Qwen3.5 weights; the original LM head supplies vocabulary logits in one forward pass. Scores for the current candidate codes are gathered and normalized within that set. The highest-scoring code identifies the selected option (B, corresponding to Label 2, in this illustration). The dashed red path denotes candidate-set cross-entropy and backpropagation to LoRA during training. Correct answers provide supervision and are excluded from the input prompt.}
\label{fig:agrijev_architecture}\end{figure*}

Figure~\ref{fig:agrijev_architecture} summarizes the AgriJev input, candidate-scoring, and adaptation workflow. Figure~\ref{fig:external_teaser} places the resulting checkpoints in an observed accuracy--cost comparison.

\subsection{Agricultural decision tasks}
A record is $r=(x,q,D,y)$: agricultural text $x$, a judgment question $q$, candidate definitions $D$, and gold index $y$. Each candidate has a name, definition, stable semantic identifier, and local position code $a_i$. Codes identify the current options; semantic identifiers retain their evaluation meaning under reordering. Single-label questions require one candidate, and binary judgments use two candidates.

\emph{Given-span entity typing} supplies an annotated mention in its original text and asks for its source type. \emph{Entity-type assertion} checks a proposed type against that annotation. \emph{Event presence} tests whether a source extraction annotation records an event category; categories are evaluated as separate binary questions. Native agricultural multiple-choice questions retain their source options and answer keys. Multi-select sources use option-level decisions where required, with reconstructed question-level exact match reported separately.

\subsection{Candidate scoring}
AgriJev uses a Qwen3.5 backbone \citep{qwen2026} and a shared LoRA adapter \citep{hu2022}. The serialized text, question, and definitions end at a fixed JSON answer prefix, with thinking disabled. Candidate codes are distinct single-token extensions at this complete boundary, even when candidate names contain several tokens. Gold labels and source metadata remain outside the model prompt.

Let $z_v$ denote the final-position logit for vocabulary token $v$, and let $a_1,\ldots,a_K$ identify the current candidate-code tokens. AgriJev returns
\begin{equation}
p_\theta(i\mid x,q,D)=\frac{\exp(z_{a_i})}{\sum_{j=1}^{K}\exp(z_{a_j})}.
\label{eq:prob}
\end{equation}
The maximum-score code is mapped back to its semantic identifier. The original language-model head supplies all required logits in one forward pass. Equation~\ref{eq:prob} defines a candidate-relative distribution; normalization does not establish calibrated confidence \citep{guo2017}.

\begin{table*}[t]\centering\footnotesize\setlength{\tabcolsep}{2pt}
\caption{Supervised internal evaluation (\%). All models use the 41,724-record training release; Qwen3.5 rows use SFT and native answer generation, while AgriJev and encoder controls score candidates. Model sizes are nominal. Family citations are given in Section~\ref{sec:models_metrics}. Typing, assertion, and event views number 3,578, 7,156, and 1,306; events are silver. BIO is a 1,173-record internal subset with fixed 27-class macro-F1. Bold marks the highest accuracy or macro-F1 in each column, including ties.}\label{tab:internal}
\begin{tabular*}{\textwidth}{@{}@{\extracolsep{\fill}}llccccc@{}}\toprule
Family & Model & Typing & Assertion & Event & BIO accuracy & BIO macro-F1\\
\midrule
\multirow{4}{*}{\shortstack[l]{AgriJev\\(ours)}} & 0.8B & 96.17 & 89.83 & 80.25 & 92.75 & 79.63\\
 & 2B & 96.70 & 95.46 & 84.00 & 94.03 & 83.77\\
 & 4B & 97.12 & 95.89 & \textbf{84.15} & 94.46 & 84.33\\
 & 9B & \textbf{97.62} & 95.30 & 81.93 & \textbf{95.57} & \textbf{86.37}\\
\midrule
\multirow{4}{*}{\shortstack[l]{Qwen3.5\\\citeyearpar{qwen2026}}} & 0.8B & 96.17 & 87.21 & 73.35 & 93.01 & 79.86\\
 & 2B & 97.07 & 95.28 & 83.46 & 94.63 & 82.75\\
 & 4B & 97.51 & \textbf{96.66} & 83.77 & 95.48 & 85.46\\
 & 9B & 97.43 & 94.37 & 81.32 & 95.31 & 85.78\\
\midrule
\multirow{2}{*}{\shortstack[l]{Other\\models}} & GLiClass-FT \citeyearpar{stepanov2025} & 96.70 & 53.51 & 43.19 & 93.52 & 82.75\\
 & RoBERTa-FT \citeyearpar{cui2020} & 92.79 & 66.24 & 47.86 & 92.75 & 82.83\\
\bottomrule\end{tabular*}
\end{table*}

\subsection{Agricultural adaptation}
AgriJev minimizes candidate-set cross-entropy,
\begin{equation}
\mathcal{L}_{\mathrm{cand}}=-z_{a_y}+\log\sum_{j=1}^{K}\exp(z_{a_j}).
\end{equation}
Training uses 41,724 records: 29,482 entity-typing examples from AgriCHN and CROP, plus 12,242 native BBK English questions. A frozen 600-record development partition validates each training condition. Candidate-order augmentation remaps the gold code through its semantic identifier. Only query/value LoRA parameters are trained; the backbone, visual modules, and vocabulary projection remain frozen. Source composition and filtering are documented in Appendix~\ref{app:data_construction}, Figure~\ref{fig:appendix_composition}, and Tables~\ref{tab:training_composition} and~\ref{tab:bbk_filtering}. Training schedules, native-SFT development losses, and loaded parameters are in Appendix~\ref{app:implementation}, Tables~\ref{tab:training_hyperparameters}, \ref{tab:native_training}, and~\ref{tab:decoder_parameter_counts}.

\subsection{Native generated-answer references}
Original Qwen3.5 models use their released weights without project fine-tuning. Qwen3.5-SFT starts from the same releases and supervises the complete answer code followed by native EOS. Its full-vocabulary response loss excludes prompt tokens. It shares the agricultural records, candidate permutations, adapter configuration, and update budget with the adapted family.

Both generated-answer references use the native chat template, disable thinking, and request only a listed option code. Greedy decoding continues until native EOS, with a maximum of 64 new tokens and a within-request KV cache. No candidate-token mask, forced one-token stop, custom assistant answer prefix, or scoring fallback is used. Unambiguous answers are mapped to semantic labels; invalid, ambiguous, or budget-exhausted outputs count as errors. Appendix~\ref{app:baselines} gives the fixed prompt and extraction rules; Appendix~\ref{app:input_template} specifies scoring-code binding.

This comparison evaluates complete adaptation and inference pipelines. Native SFT and AgriJev also differ in prompt boundary and supervised response, so their comparison does not isolate training normalization. Prior same-inference scoring controls address that narrower question (Appendix~\ref{app:scoring_controls}; Tables~\ref{tab:adaptation_ablation}, \ref{tab:old_internal}, \ref{tab:old_external_scales}, and~\ref{tab:old_supervised_timings}) and show competitive answer-token SFT.

\section{Evaluation}
We assess agricultural training through held-out internal judgments and transfer through external views without target-dataset training. Original-context groups are separated before deriving internal task records. The three internal tasks reuse source annotations, with event labels explicitly marked as silver. BIO is an internal typing subset, not additional external evidence.

\subsection{Data sources and partitions}\label{sec:data_sources}
The internal test contains 12,040 views over 1,505 original groups: 3,578 typing, 7,156 assertion, and 1,306 event views. Typing and assertion share labels and groups. Table~\ref{tab:internal_inventory} documents these views. The BIO subset contains 1,173 records over 251 groups with a fixed 27-class inventory. Training includes typing and native English choices; assertion and event tests assess additional judgments without direct task supervision.

Source-derived judgments use AgriCHN annotations \citep{agrichncard} and CROP extraction resources \citep{zhang2024,crop_resources}. Added BBK English choices are training-only \citep{devane2025}. We explicitly repurpose that resource's released English test partition and exclude BBK from external evaluation. Appendix~\ref{app:data_construction} records filtering, overlap exclusions, source licenses, and the frozen release.

External knowledge views use AgriEval \citep{yan2026}, CROP \citep{zhang2024,crop_resources}, CMMLU agronomy \citep{li2023cmmlu}, and C-Eval plant protection and veterinary medicine \citep{huang2023ceval}. They also include three TMMLU+ subjects \citep{tam2024tmmlu}, five SeedBench views \citep{ying2025seedbench}, and DEW-LogiQ water-management choices \citep{dewlogiqcard}. Text judgments use ORKG AgriNER \citep{dsouza2024agriner}, FAIRagro NER \citep{abdelmalak2025fairagro}, and released Ag-valuate judged pairs \citep{koopman2022agask}. There are 18 prepared external views, including paired CROP reordering; only 17 enter the main mean. These views are not 17 independently created datasets.

\subsection{Models and metrics}\label{sec:models_metrics}
The primary references are original Qwen3.5 and native-answer Qwen3.5-SFT at 0.8B, 2B, 4B, and 9B. Complementary references include AgriReason-4B \citep{agrireasoncard}, Akshara-7B \citep{cropin2024akshara}, GLiClass \citep{stepanov2025}, NLI \citep{laurer2023universal}, BGE \citep{xiao2024cpack}, and cached Jev 1.13.0 choices \citep{typesafejev,almeida2026jev}. Their declared scoring or service contracts remain distinct. We also include GLiClass-FT and RoBERTa-FT \citep{cui2020} as supervised internal controls. Appendix~\ref{app:baselines} specifies model interfaces and coverage; BGE's partial reading-view results are explicitly marked.

We report accuracy separately by task and view. BIO macro-F1 uses the same 27 semantic classes and zero division set to zero. The external mean weights all 17 complete views equally, rather than pooling transformed records. Table~\ref{tab:external} summarizes SeedBench by its five-view mean; Appendix~\ref{app:external_breakdown} retains every view and reconstruction metric. Original and reordered CROP share questions and are not treated as independent tests. Full input qualification prohibits silent truncation. Detailed training and measurement settings remain in the appendix.

\section{Results and Analysis}
\subsection{Supervised internal evaluation}
Table~\ref{tab:internal} reports completed internal evaluations after agricultural training. AgriJev-9B reached 97.62\% typing accuracy and 86.37\% BIO macro-F1. Its 4B counterpart had higher assertion and event accuracy, at 95.89\% and 84.15\%. GLiClass-FT and RoBERTa-FT were competitive on typing, but weaker on the additional assertion and event judgments.

Native-answer SFT also performed strongly. Its 4B model reached 96.66\% assertion accuracy and 85.46\% BIO macro-F1, compared with AgriJev-4B's 95.89\% and 84.33\%. AgriJev-9B had higher BIO macro-F1 than the 9B SFT generator, at 86.37\% versus 85.78\%. These task-level outcomes expose deployment differences without uniform internal superiority. Appendix~\ref{app:bio_references} provides off-the-shelf and class-level BIO results (Tables~\ref{tab:bio_unadapted} and~\ref{tab:bio_per_class}).

\subsection{Zero-shot external transfer}
\begin{table*}[t]\centering\footnotesize\setlength{\tabcolsep}{2pt}
\caption{Zero-shot external accuracy (\%). Qwen3.5, Qwen3.5-SFT, and AgriJev use 9B; AgriReason uses 4B, Akshara 7B, and Jev version 1.13.0. Both Qwen conditions generate answers; AgriJev scores candidates. Header citations identify model families; SFT follows Section~\ref{sec:models_metrics}. Dataset citations are in Section~\ref{sec:data_sources}. SeedBench averages five view accuracies equally; $\dagger$ marks BGE averages using qualified subsets for three reading views, detailed in Appendix~\ref{app:baselines}. Bold marks only the highest complete-coverage result in each row, including ties. Paired CROP reordering and full view results appear in Appendix~\ref{app:external_breakdown}.}\label{tab:external}
\begin{tabular*}{\textwidth}{@{}@{\extracolsep{\fill}}lccccccccc@{}}\toprule
Dataset & \shortstack{Jev\\\citeyearpar{typesafejev}} & \shortstack{AgriJev\\(ours)} & \shortstack{Qwen3.5\\\citeyearpar{qwen2026}} & \shortstack{Qwen3.5-SFT\\\citeyearpar{qwen2026}} & \shortstack{AgriReason\\\citeyearpar{agrireasoncard}} & \shortstack{GLiClass\\\citeyearpar{stepanov2025}} & \shortstack{NLI\\\citeyearpar{laurer2023universal}} & \shortstack{BGE\\\citeyearpar{xiao2024cpack}} & \shortstack{Akshara\\\citeyearpar{cropin2024akshara}}\\
\midrule
AgriEval & \textbf{67.60} & 62.20 & 58.18 & 62.30 & 37.29 & 26.21 & 40.23 & 43.02 & 42.49\\
CROP & \textbf{93.52} & 91.10 & 90.18 & 93.02 & 45.19 & 19.31 & 42.38 & 28.40 & 66.53\\
CMMLU agronomy & \textbf{87.57} & 76.33 & 72.19 & 76.92 & 28.40 & 26.63 & 23.67 & 30.77 & 38.46\\
TMMLU+ agriculture & \textbf{76.39} & 65.97 & 52.78 & 61.81 & 23.61 & 29.86 & 22.22 & 22.22 & 27.78\\
C-Eval plant protection & \textbf{90.45} & 83.42 & 80.40 & 82.91 & 43.22 & 30.15 & 27.14 & 25.13 & 38.69\\
C-Eval veterinary & \textbf{92.38} & 91.43 & 83.81 & 90.48 & 40.48 & 30.48 & 27.14 & 29.52 & 34.76\\
TMMLU+ pathology & \textbf{79.15} & 65.02 & 60.78 & 63.60 & 31.45 & 22.97 & 24.38 & 28.62 & 30.74\\
TMMLU+ pharmacology & \textbf{92.01} & 82.71 & 73.98 & 77.88 & 31.41 & 23.98 & 24.54 & 24.91 & 34.57\\
SeedBench (5-view mean) & \textbf{83.25} & 73.41 & 67.44 & 63.41 & 45.69 & 48.35 & 43.74 & 37.62$^\dagger$ & 56.51\\
DEW-LogiQ & \textbf{73.03} & 63.20 & 65.45 & 62.64 & 32.87 & 33.43 & 32.30 & 22.47 & 46.35\\
AgriNER & \textbf{61.95} & 61.25 & 53.25 & 55.92 & 37.24 & 35.61 & 19.03 & 23.90 & 40.84\\
FAIRagro & \textbf{92.95} & 89.42 & 90.71 & 87.50 & 43.91 & 17.63 & 19.23 & 15.71 & 54.49\\
AgVaLuate subset & 71.33 & 83.33 & 63.33 & \textbf{86.00} & 40.00 & 84.00 & 84.00 & 5.33 & 66.67\\
\bottomrule\end{tabular*}
\end{table*}

Table~\ref{tab:external} reports external results, while Figure~\ref{fig:external_teaser} shows scale comparisons. Complete external and encoder results appear in Tables~\ref{tab:external_full} and~\ref{tab:external_encoders}; Tables~\ref{tab:external_small_scales} and~\ref{tab:external_large_scales} retain every scale. AgriJev-9B reached 75.44\% mean accuracy across 17 views, versus 69.54\% for original generation and 71.65\% for SFT generation. The differences were 5.89 and 3.79 percentage points, calculated before rounding. It exceeded the original generator on 13 views and the SFT generator on 11.

The remaining scales showed smaller differences against SFT generation. AgriJev-2B and AgriJev-4B reached 59.19\% and 70.47\%, exceeding their SFT means by 0.27 and 0.57 percentage points. At 0.8B, AgriJev's 53.55\% was 0.07 points below SFT's 53.62\%. These descriptive point estimates do not establish equivalence or statistical significance.

Differences were task-dependent. On CROP, the 9B SFT generator reached 93.02\%, exceeding AgriJev-9B's 91.10\%. On SeedBench 3-2, AgriJev-9B reached 75.50\%, versus 55.57\% for original generation and 39.98\% for SFT generation. The generators produced 113 and 236 invalid codes among 808 decisions; these outputs remained in the denominator. Table~\ref{tab:generation_invalid} records invalid-output counts (Appendix~\ref{app:external_breakdown}); Table~\ref{tab:multiselect_generation} reports separate question-level reconstruction.

Jev 1.13.0 reached 82.04\% on the same descriptive mean. Agricultural adaptation narrowed the original generator's 12.49-point gap to 6.60 points. Jev remained stronger on several knowledge views, while AgriJev had higher passage-relevance accuracy.

\subsection{Inference costs}
\begin{figure}[t]\centering
\includegraphics[width=\columnwidth]{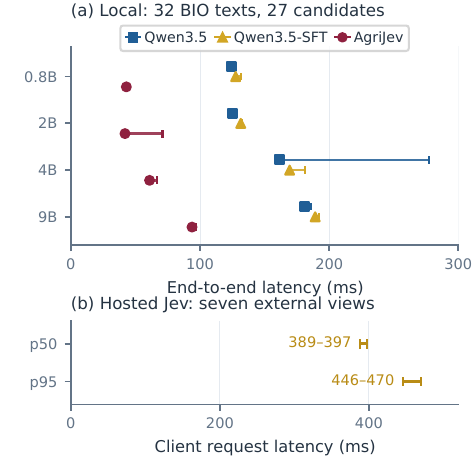}
\caption{\textbf{Observed deployment latency.} (a) Points give p50 and right-hand caps p95 on 32 BIO texts with 27 candidates, A100 40GB, batch 1, and 96 timings per model. Both Qwen families generate answers; AgriJev scores candidates. (b) Hosted Jev ranges summarize seven external views and include service, network, waiting, and retries. Panels have different inputs and timing conditions.}
\label{fig:inference_cost}\end{figure}

Figure~\ref{fig:inference_cost} compares actual native generation with candidate scoring. On the fixed local workload, AgriJev's 0.8B, 2B, 4B, and 9B medians were 42.96, 41.90, 60.93, and 93.80\,ms. Original generated-answer medians were 124.04, 125.17, 161.79, and 181.24\,ms. The observed generation/scoring ratios ranged from 1.93 to 2.99 across scales. At 9B, SFT generation took 189.16\,ms, giving a ratio of 2.02 relative to AgriJev.

Inputs and hardware class were fixed, while native prompt boundaries and outputs followed each method's contract. Generation returned a decoded answer; scoring returned a semantic decision with candidate-relative probabilities. The measurements include the specified preparation, transfers, synchronized model execution, and native result conversion. They exclude loading and file I/O, and do not isolate training as the cause of latency differences. Appendix~\ref{app:cost} and Table~\ref{tab:current_timings} record complete timings, cache, and implementation conditions.

Hosted Jev had per-view p50 values of 388.53--397.24\,ms and p95 values of 446.08--469.79\,ms across 22,162 new requests. Those client-side durations include remote-service overhead and use different inputs. They describe the observed service condition rather than a hardware-matched speed comparison.

\section{Discussion}\label{sec:discussion}
\begin{figure}[t]\centering
\includegraphics[width=\columnwidth]{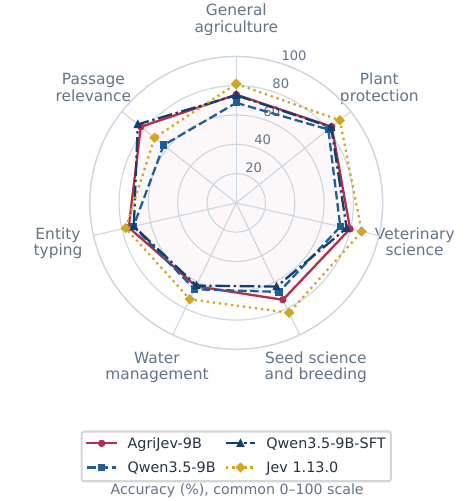}
\caption{\textbf{External capability profiles.} Seven source-defined blocks contain 4, 1, 3, 5, 1, 2, and 1 views. Each axis averages complete view accuracies on the same 0--100\% scale. Qwen3.5 and Qwen3.5-SFT use native generation; AgriJev uses candidate scoring; Jev uses hosted choices. Block definitions appear below and in Appendix~\ref{app:capability_profiles}. Paired CROP reordering is excluded.}
\label{fig:capability_radar}\end{figure}
\begin{figure*}[t]\centering
\includegraphics[width=\textwidth]{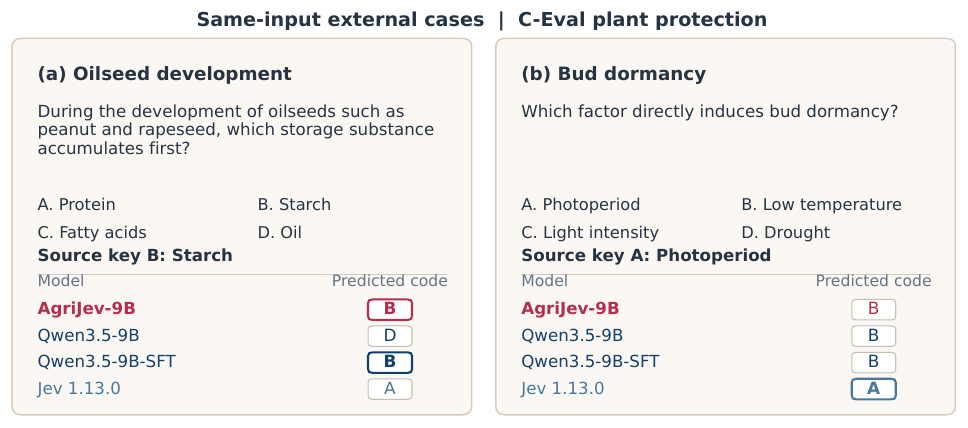}
\caption{\textbf{Actual decisions on two external questions.} Original Chinese inputs are translated only for display. Rows retain saved AgriJev, original-Qwen generation, SFT generation, and Jev predictions on identical IDs. Outlined codes match the source key. No confidence is inferred for generated answers. The original two selected questions illustrate successes and failures; they do not estimate their frequency.}
\label{fig:heldout_case}\end{figure*}

\paragraph{Capabilities depend on the source task.}
Figure~\ref{fig:capability_radar} separates overall accuracy from its source-defined capability profile. AgriJev-9B improved on original generation in six of seven block averages, with water management as the exception. Against SFT generation, it improved in six blocks, with passage relevance as the exception. Individual views still differed within those blocks, including the stronger SFT result on CROP.

General agriculture combines AgriEval, CROP, CMMLU agronomy, and TMMLU+ agriculture. Plant protection uses C-Eval plant protection; veterinary science combines C-Eval veterinary medicine and two TMMLU+ veterinary subjects. Seed science and breeding contains five SeedBench views. Water management uses DEW-LogiQ, entity typing combines AgriNER and FAIRagro, and passage relevance uses the AgVaLuate judged subset. These are reporting groups based on source purpose, not disjoint subject annotations for every question.

\paragraph{Selection and answer construction have different failure modes.}
Candidate scoring always selects from the supplied inventory, whereas native generators can emit an unlisted or unparseable answer. The SeedBench failures show that successful format compliance on internal tasks need not transfer to every external decision formulation. Restricting the final decision to candidates removes this output-validity failure, but does not ensure the selected semantic label is correct. Scores likewise remain conditional on the input and inventory, without demonstrated probability calibration.

The main comparison concerns complete pipelines, including adaptation and inference. We credit prior selection and label-binding work, and attribute the present contribution to agricultural preparation, adapted weights, and their measured deployment tradeoffs.

\paragraph{Same-input qualitative cases.}
Figure~\ref{fig:heldout_case} updates the two previously selected C-Eval questions with actual generated answers. On oilseed development, AgriJev and SFT generation selected B, while original generation selected D and Jev selected A. On bud dormancy, all three local models selected B, while Jev selected the source key A. Appendix~\ref{app:case_audit} records provenance and reproducibility boundaries. These disagreements illustrate complementary failures without establishing error prevalence or internal reasoning mechanisms.

\paragraph{Scale and deployment requirements.}
Increasing scale did not improve every task. AgriJev-9B had higher typing and BIO scores than 4B, but lower assertion and event accuracy. AgriJev-4B also exceeded 9B on CROP, at 91.75\% versus 91.10\%, while its measured median latency was lower. The smaller checkpoints provided faster scoring with lower suite accuracy. Choosing weights therefore requires the task-level evidence as well as aggregate accuracy and the relevant hardware workload.

\paragraph{Answer validity and semantic accuracy.}
Invalid outputs explain part of the generation gap, but they do not account for the entire 9B mean difference. Consider an arithmetic upper bound that makes every invalid answer correct, preserving all valid decisions and the fixed view weights. Original generation would rise from 69.54\% to 71.02\%; SFT generation would rise from 71.65\% to 74.15\%. AgriJev's observed 75.44\% remains higher, by 4.41 and 1.29 percentage points. Table~\ref{tab:generation_invalid} specifies this bound, which assumes unavailable gold answers and is not an achieved repair result. The comparison therefore involves semantic decision differences as well as output validity. Prompt boundaries, supervised responses, and inference contracts still differ, so this diagnostic does not isolate loss normalization as their cause.

\paragraph{Interpreting the size--latency tradeoff.}
The nearly equal bubble areas within each scale in Figure~\ref{fig:external_teaser} make the parameter budget explicit. AgriJev retains the Qwen3.5 backbone and vocabulary projection, with a small adapter overhead. Its lower measured latency therefore does not indicate model compression. The gain concerns complete inference contracts: a candidate-relative decision from one forward call versus constructing and extracting a native answer sequence. Both conditions retain input preparation and result conversion in their measurement boundaries. Applications that need a decision over explicit candidates can use this interface directly. Applications requiring explanations or long-form responses need a different output contract, whose cost and accuracy are not measured here. Replication, calibration, and new candidate inventories remain necessary tests before broader deployment claims.

\section{Conclusion}
\label{sec:conclusion}
AgriJev connects source-preserving agricultural evaluation, a shared candidate-scoring interface, and adapted Qwen3.5 weights at four scales. It returns candidate scores and relative probabilities in one forward call. AgriJev-9B reached 75.44\% mean accuracy across 17 external views, exceeding original and SFT answer generation by 5.89 and 3.79 percentage points. Its median latency was approximately half that of either same-scale generator on the fixed A100 workload. Task-level reversals and competitive prior scoring controls bound these findings. The contribution is a documented agricultural deployment choice, with accuracy and cost evidence; candidate-set training does not guarantee universal superiority or calibrated confidence.
\label{sec:main_end}

\section*{Limitations}
Typing and assertion reuse source annotations; BIO is an internal subset. Generated extraction labels and closed-world event negatives remain silver, without additional expert adjudication. External views differ in measurement units, and paired CROP views reuse questions. Option accuracy is distinct from question-level exact match. Unseen-label transfer, independent calibration, and unrestricted agricultural reasoning have not been tested. Native generators disable thinking and request short codes, so timing does not characterize long-form reasoning. The 32-input workload does not guarantee speed on other hardware, batches, or candidate inventories. Single-run details appear in Appendix~\ref{app:implementation}. Public hosting and final artifact-access arrangements remain incomplete.

\section*{Ethical Considerations}
We preserve source attribution, original answer keys, annotation-quality metadata, and applicable access conditions. BBK's released English test partition is used only for training and excluded from external evaluation; Appendix~\ref{app:data_construction} records filtering and CC BY 4.0 attribution. Derived judgments retain their source status. Candidate-relative probabilities are not calibrated disease risks or substitutes for expert advice. Invalid outputs remain errors without dropped records. Public redistribution of processed data or model artifacts is not claimed.

\FloatBarrier

\clearpage\appendix
\raggedbottom
\setcounter{dbltopnumber}{3}
\makeatletter\setlength{\@fptop}{0pt}\setlength{\@fpsep}{12pt}\setlength{\@fpbot}{0pt plus 1fil}
\setlength{\@dblfptop}{0pt}\setlength{\@dblfpsep}{12pt}\setlength{\@dblfpbot}{0pt plus 1fil}\makeatother
\section{Data Sources, Prepared Views, and Split Protocol}\label{app:data_construction}
\begin{figure*}[t]\centering
\includegraphics[width=\textwidth]{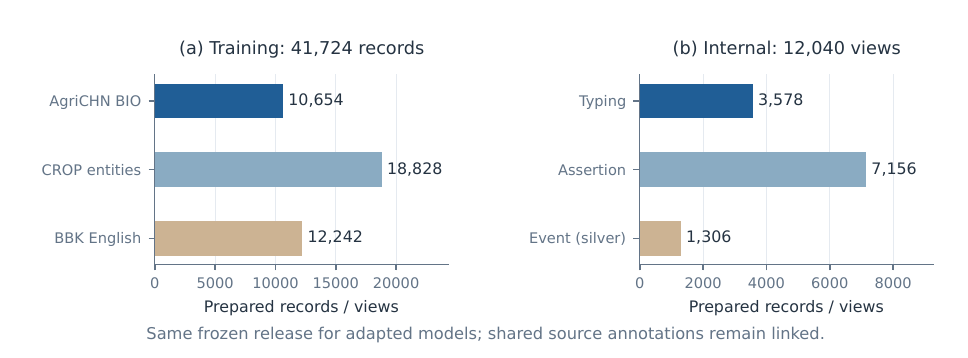}
\caption{\textbf{Prepared-release composition.} (a) The shared 41,724-record training release combines 10,654 AgriCHN source-BIO examples, 18,828 CROP-generated entity examples, and 12,242 native BBK English choices. (b) The 12,040 internal views contain typing, assertion, and event tasks. Counts are exact frozen-release totals, not independent annotation counts: typing and assertions share source labels, and BIO is a typing subset.}
\label{fig:appendix_composition}\end{figure*}

This study prepares candidate-decision records from existing resources and their released annotations. The training release and evaluation suite retain source attribution, semantic labels, original groups, and annotation quality; the conversions do not constitute a new independently expert-annotated agricultural corpus. Figure~\ref{fig:appendix_composition} and Table~\ref{tab:training_composition} document the current release.

\paragraph{Source provenance and training composition}
The frozen training release contains 41,724 records: 10,654 AgriCHN source-BIO entity-typing examples \citep{agrichncard}, 18,828 CROP-generated entity-typing examples \citep{zhang2024,crop_resources}, and 12,242 native English BhashaBench-Krishi (BBK) questions \citep{devane2025}. Thus the 29,482-record entity component combines AgriCHN and CROP rather than AgriCHN alone. Its language counts are 21,950 Chinese and 7,532 English examples. BBK contributes four-option English questions, with no synthesized distractors or rationales. The source BBK release calls its partition test; we repurpose accepted English records for training and exclude BBK from external evaluation. The official BBK card lists CC BY 4.0 and access conditions; redistribution must retain source attribution and applicable access requirements.

\begin{table}[t]\centering\footnotesize\setlength{\tabcolsep}{2pt}
\caption{Prepared training records and preserved source-group IDs. Group counts identify provenance units within each source, not newly independent annotations. CROP entity labels remain generated supervision.}
\label{tab:training_composition}
\begin{tabular*}{\columnwidth}{@{}@{\extracolsep{\fill}}lrrl@{}}\toprule
Source & Records & Groups & Supervision\\\midrule
AgriCHN & 10,654 & 2,135 & Source BIO\\
CROP entities & 18,828 & 4,853 & Generated labels\\
BBK English & 12,242 & 12,111 & Native answers\\
Total & 41,724 & \multicolumn{2}{c}{Two training task families}\\
\bottomrule\end{tabular*}\end{table}
\paragraph{Cleaning and group isolation}
The entity component comes from the frozen original training partition. Original text and near-duplicate context groups are separated before task views are derived. The frozen development partition supplies 600 records for development validation and contains no BBK questions. Of 12,648 BBK English rows, 405 are quarantined: 259 duplicate-option rows, three empty-option rows, and 143 duplicate-question rows. One further exact overlap with the external suite is excluded, leaving 12,242. The final training file has SHA256 \nolinkurl{6ea1948880102d1ce883850d1bd84e2dc9d89fed7a6557452882cda3a07676aa}. The entity-source partition audit found zero original-group and exact-context overlap with development and internal test; the BBK guard checks external question overlap. Format and overlap qualification are distinct from expert label adjudication.

\begin{table}[t]\centering\footnotesize\setlength{\tabcolsep}{2pt}
\caption{BBK English filtering, applied before adding native choices to the training release. Each exclusion count is recorded by the frozen preparation audit.}
\label{tab:bbk_filtering}
\begin{tabular*}{\columnwidth}{@{}@{\extracolsep{\fill}}lr@{}}\toprule
Stage & Rows\\\midrule
Released English rows & 12,648\\
Duplicate-option exclusion & 259\\
Empty-option exclusion & 3\\
Duplicate-question exclusion & 143\\
External exact-overlap exclusion & 1\\
Accepted training records & 12,242\\
\bottomrule\end{tabular*}\end{table}
\paragraph{Task conversions and supervision}
Each record preserves state, question (with an instructions alias where needed), candidate identifiers, names and definitions, semantic target, source, original group, and split. Given-span typing retains an annotated mention in its original text. Type assertions derive positive and negative judgments from that same annotation. Event targets follow source extraction lists, with closed-world negatives marked silver. Gold fields and provenance metadata are excluded from model input. Candidate shuffling recomputes answer positions through semantic identifiers; positional codes are not treated as ontology classes.

Training uses entity typing and native four-option English questions. Assertions and event presence are additional source-derived evaluation tasks, without direct supervision in this training release.

\paragraph{Frozen internal task views}
The frozen test contains 12,040 task views over 1,505 original groups: 3,578 typing views over 877 groups, 7,156 assertions over the same 877 groups, and 1,306 event views over 638 groups. Group counts across tasks are not additive. Source-BIO typing is a subset of 1,173 records over 251 groups with a fixed 27-class inventory. The remaining typing sources include generated extraction labels. These task transformations reuse source annotations rather than constituting three independently annotated datasets.

\begin{table}[t]\centering\footnotesize\setlength{\tabcolsep}{2pt}
\caption{Internal task-view inventory. BIO is a subset of typing. Typing and assertion group counts overlap; counts across these rows must not be added to infer independent groups.}
\label{tab:internal_inventory}
\begin{tabular*}{\columnwidth}{@{}@{\extracolsep{\fill}}lrr@{}}\toprule
Task view & Records & Groups\\\midrule
Entity typing & 3,578 & 877\\
Type assertion & 7,156 & 877\\
Event presence & 1,306 & 638\\
All internal views & 12,040 & 1,505\\
BIO subset & 1,173 & 251\\
\bottomrule\end{tabular*}\end{table}
\paragraph{Frozen external views}
The 18 views and their record counts appear in Table~\ref{tab:external_full}. Native CROP has 5,043 questions over 5,033 groups after two rows with normalized duplicate options are quarantined. Its reordered view retains the same questions and remapped answers. The adaptation data exclude open CROP QA because 3,994 normalized stems overlap the benchmark. Possible backbone pretraining exposure is not resolved by adaptation overlap checks. AgVaLuate is the released judged subset; no larger adjudicated cohort is claimed. Multi-select tasks are represented by option-level decisions where required, so transformed record counts differ from source-question counts. Frozen manifests retain source versions, original row IDs, semantic labels, group IDs, and evaluation-file hashes.

\paragraph{Qualification and annotation boundaries}
Input qualification checks candidate validity, semantic-gold consistency, complete input lengths, source-group separation, and recorded overlap exclusions. These checks verify format and partition integrity. No new expert adjudication of the converted records has been completed. Source BIO labels, generated extraction labels, and native answer keys remain distinguishable; closed-world event negatives are silver. Open CROP QA is excluded from the current adaptation release because its normalized stems overlap the external CROP questions. The BBK overlap guard excludes one accepted-format question matching an external reading view.

The three source resources retain their own attribution and access conditions. AgriCHN and BBK cards report CC BY 4.0; the CROP resources report CC BY-NC 4.0. Repackaging and model training do not erase those source conditions. This appendix documents the materials actually used, without claiming that all processed artifacts or trained weights have already been publicly hosted.

\FloatBarrier
\section{Model Training and Checkpoints}\label{app:implementation}

AgriJev and native-answer Qwen3.5-SFT start from the same original Qwen3.5 releases with fresh query/value LoRA adapters. Both use the 41,724-record release, frozen 600-record development partition, seed 13, one epoch, effective batch 16, and 2,608 updates. The decoder checkpoint is the end of the fixed epoch; development loss is reported for validation, and intermediate recovery checkpoints are ineligible for selection. No test metric selects a checkpoint. Table~\ref{tab:training_hyperparameters} gives the shared schedule.
\begin{table}[t]\centering\footnotesize\setlength{\tabcolsep}{2pt}
\caption{Frozen training settings. AgriJev and native-answer SFT share this decoder schedule; their supervised outputs differ. Encoder controls share data and update count but use a different optimizer learning rate and checkpoint-selection schedule.}
\label{tab:training_hyperparameters}
\begin{tabular*}{\columnwidth}{@{}@{\extracolsep{\fill}}lll@{}}\toprule
Setting & Decoders & Encoders\\\midrule
Training records & 41,724 & 41,724\\
Development records & 600 & 600\\
Epochs / updates & 1 / 2,608 & 1 / 2,608\\
Random seed & 13 & 13\\
Effective batch & 16 & 16\\
Learning rate & $10^{-4}$ & $2\times10^{-5}$\\
Training GPUs & 4 A100s & 1 A100\\
Per-GPU microbatch & 1 record & 1 record\\
Accumulation & 4 records/rank & 16 records\\
LoRA rank / alpha & 16 / 32 & Full fine-tuning\\
LoRA dropout & 0.05 & Not applicable\\
Backbone gradients & Frozen & Enabled\\
\bottomrule\end{tabular*}\end{table}
\paragraph{Native-answer SFT objective.}
The response contains a local code and native EOS, both supervised by full-vocabulary cross-entropy; prompt tokens have zero loss weight. Loss is averaged over the supervised response tokens within each record. Candidate permutations and gold remapping are shared with AgriJev. Complete code-plus-EOS supervision covers two tokens per development record (1,200 tokens over 600 records). Teacher-forced development CE in Table~\ref{tab:native_training} is not generated-answer accuracy. At inference, the native template may emit formatting or boundary tokens in addition to the code; the measured timing requests averaged four generated tokens. The native response objective and chat boundary differ from AgriJev's single-position candidate-set loss, so the comparison evaluates complete pipelines, rather than a loss-only ablation.
\begin{table}[t]\centering\footnotesize\setlength{\tabcolsep}{2pt}
\caption{Completed native-answer SFT runs. Minutes denote recorded training time; development CE is teacher-forced response loss at the fixed-epoch checkpoint. These timings are not inference latency.}\label{tab:native_training}
\begin{tabular*}{\columnwidth}{@{}@{\extracolsep{\fill}}lcc@{}}\toprule
Scale & Minutes & Dev. response CE\\
\midrule
0.8B & 64.42 & 0.2160\\
2B & 37.59 & 0.1178\\
4B & 89.45 & 0.1074\\
9B & 49.64 & 0.1246\\
\bottomrule\end{tabular*}
\end{table}
\paragraph{Supervised encoder controls.}
GLiClass-FT starts from \nolinkurl{knowledgator/gliclass-multilang-mini}, revision \nolinkurl{0bd888b6c3ef9fca5f0a9d407bddfbbc7623486b}. RoBERTa-FT starts from \nolinkurl{hfl/chinese-roberta-wwm-ext}, revision \nolinkurl{5c58d0b8ec1d9014354d691c538661bf00bfdb44}, and scores each candidate with a scalar CLS cross-encoder head. Both use all 41,724 training records, the same 600 development records, seed 13, one epoch, effective batch 16, 2,608 updates, and learning rate $2\times10^{-5}$. All encoder parameters are trainable. Checkpoints are selected by minimum development candidate NLL at updates 1,000, 2,000, and 2,608; this differs from the decoder fixed-epoch rule. GLiClass retains full inputs under a 4,096-token budget. RoBERTa uses complete 512-token windows over the full query/candidate pair, at most three windows per training candidate, combined by token-pair-weighted mean, without discarding input tokens. Both use FP32 inference without candidate-embedding caching. The encoder comparisons share supervision and update count, but differ in architecture, optimization, and checkpoint selection.

\begin{table}[t]\centering\footnotesize\setlength{\tabcolsep}{2pt}
\caption{Actual loaded and trainable parameter counts for each decoder scale. AgriJev and native-answer SFT at the same scale share these counts; nominal size labels are not substituted for the measured totals.}
\label{tab:decoder_parameter_counts}
\begin{tabular*}{\columnwidth}{@{}@{\extracolsep{\fill}}lrr@{}}\toprule
Scale & Loaded & Trainable\\\midrule
0.8B & 853,624,896 & 638,976\\
2B & 2,214,077,248 & 835,584\\
4B & 4,541,100,544 & 1,835,008\\
9B & 9,412,041,968 & 2,228,224\\
\bottomrule\end{tabular*}\end{table}
\paragraph{Revisions and replication.}
The 0.8B, 2B, 4B, and 9B base revisions are respectively \nolinkurl{2fc06364715b967f1860aea9cf38778875588b17}, \nolinkurl{15852e8c16360a2fea060d615a32b45270f8a8fc}, \nolinkurl{851bf6e806efd8d0a36b00ddf55e13ccb7b8cd0a}, and \nolinkurl{c202236235762e1c871ad0ccb60c8ee5ba337b9a}. Table~\ref{tab:decoder_parameter_counts} measures loaded backbone-plus-adapter parameters; released original models exclude adapter parameters. Each adapted condition has one completed seed-13 run. Multi-seed replication has not been completed, and descriptive differences do not establish significance. External view means and SeedBench means average views, not seeds.

\paragraph{Optimization and complete inputs.}
Decoder AdamW uses learning rate $10^{-4}$, $\beta_1=0.9$, $\beta_2=0.999$, $\epsilon=10^{-8}$, weight decay 0.01, and clipping at 1.0. Four A100 ranks each use microbatch 1 and four accumulation steps. The last update contains 12 real records; distributed padding has zero loss weight. LoRA rank is 16, alpha 32, and dropout 0.05, targeting query/value projections in full-attention layers. Backbone, visual modules, and the vocabulary projection remain frozen. Backbone parameters are BF16, adapters and decoder losses FP32. Recorded software is PyTorch 2.6.0+cu124, Transformers 5.2.0, and PEFT 0.18.1.

Training uses a 2,048-token complete-input budget. Under the scoring serialization, train and development maxima are 971 and 857 tokens. The native-generation template is qualified independently: its longest evaluated input is 1,723 tokens, under a 32,768-token qualification limit, with zero truncated inputs at every scale. Lengths from one prompt are not substituted for the other. Encoder updates normalize gradients over 16 records (12 at the last update); parameters are FP32 with BF16 training autocast. RoBERTa windows run in chunks of eight, with weight decay excluded for bias and scalar/vector parameters.

\FloatBarrier
\section{Internal BIO References and Class-Level Results}\label{app:bio_references}

BIO is an internal subset of 1,173 records over 251 groups. Table~\ref{tab:bio_unadapted} distinguishes native generated-answer references from the prior candidate-scoring references. None of these off-the-shelf checkpoints is trained on the project's agricultural release. Accuracy is record-weighted; macro-F1 averages the fixed 27 semantic classes with zero division set to zero. Answer-code positions are not semantic classes.
\begin{table}[t]\centering\footnotesize\setlength{\tabcolsep}{2pt}
\caption{Off-the-shelf BIO references (\%). Generate denotes native answer generation; Score retains the prior candidate-scoring contract. Identical source records and the fixed 27-class metric are used; the two contracts are not interchangeable.}\label{tab:bio_unadapted}
\begin{tabular*}{\columnwidth}{@{}@{\extracolsep{\fill}}lccc@{}}\toprule
Model & Mode & Accuracy & Macro-F1\\
\midrule
Qwen3.5-0.8B & Generate & 31.12 & 22.67\\
Qwen3.5-2B & Generate & 52.00 & 39.25\\
Qwen3.5-4B & Generate & 80.82 & 70.35\\
Qwen3.5-9B & Generate & 86.45 & 78.43\\
Qwen3.5-0.8B & Score & 38.02 & 29.58\\
Qwen3.5-2B & Score & 66.07 & 57.31\\
Qwen3.5-4B & Score & 85.85 & 76.54\\
Qwen3.5-9B & Score & 87.55 & 80.07\\
AgriReason-4B & Score & 34.61 & 23.19\\
Akshara-7B & Score & 55.41 & 49.48\\
GLiClass & Score & 31.12 & 26.09\\
NLI & Score & 33.50 & 27.62\\
BGE & Score & 24.98 & 16.65\\
\bottomrule\end{tabular*}
\end{table}
\paragraph{Fixed-inventory decomposition.} English glosses in Table~\ref{tab:bio_per_class} retain the source ontology. Two classes have zero gold support and remain in the macro-F1 denominator. Prior token-SFT scores are included for continuity, under a distinct scoring contract.
\begin{table*}[t]\centering\small\setlength{\tabcolsep}{2pt}
\caption{BIO class-level F1 (\%) at 9B. Qwen and native SFT generate answers; AgriJev and prior token-SFT score candidates. Support counts gold records. English names are display glosses, not relabeled classes.}\label{tab:bio_per_class}
\begin{tabular*}{\textwidth}{@{}@{\extracolsep{\fill}}lccccc@{}}\toprule
Original class (English gloss) & Support & Qwen gen. & SFT gen. & AgriJev & Token-SFT\\
\midrule
Time & 52 & 93.33 & 94.34 & 95.24 & 95.24\\
Soil moisture & 9 & 94.74 & 94.74 & 100.00 & 100.00\\
Aquatic products & 4 & 100.00 & 100.00 & 100.00 & 100.00\\
Fruits & 40 & 89.89 & 100.00 & 100.00 & 96.20\\
Diseases, pests, weeds & 203 & 85.71 & 98.26 & 98.52 & 98.52\\
Vegetables & 59 & 84.38 & 92.19 & 89.39 & 90.08\\
Organizations & 5 & 100.00 & 100.00 & 100.00 & 100.00\\
Hydrological phenomena & 0 & 0.00 & 0.00 & 0.00 & 0.00\\
Products & 4 & 28.57 & 66.67 & 66.67 & 66.67\\
Agricultural facilities & 8 & 88.89 & 82.35 & 88.89 & 82.35\\
Nutrients & 43 & 91.30 & 96.63 & 97.73 & 97.73\\
Person names & 5 & 90.91 & 90.91 & 90.91 & 90.91\\
Weather phenomena & 55 & 91.89 & 91.43 & 96.36 & 96.36\\
Temperature & 15 & 93.75 & 85.71 & 93.75 & 93.75\\
Cash crops & 68 & 66.67 & 85.93 & 84.38 & 84.85\\
Agricultural equipment & 9 & 80.00 & 80.00 & 80.00 & 80.00\\
Agricultural techniques & 159 & 93.81 & 97.81 & 98.42 & 98.43\\
Livestock / poultry & 4 & 100.00 & 100.00 & 100.00 & 85.71\\
Soil types & 12 & 100.00 & 96.00 & 96.00 & 100.00\\
Water areas & 4 & 66.67 & 100.00 & 100.00 & 100.00\\
Pesticides & 76 & 93.88 & 96.64 & 96.64 & 96.64\\
Fertilizers & 96 & 90.82 & 96.88 & 96.88 & 96.34\\
Fodder crops & 11 & 60.00 & 86.96 & 80.00 & 83.33\\
Diseases & 0 & 0.00 & 0.00 & 0.00 & 0.00\\
Grain crops & 70 & 83.78 & 94.96 & 92.09 & 94.89\\
Other organisms & 47 & 51.28 & 88.64 & 90.91 & 91.11\\
Place names & 115 & 97.32 & 99.12 & 99.12 & 99.56\\
\bottomrule\end{tabular*}
\end{table*}

\section{Baseline and Evaluation Protocols}\label{app:baselines}

\paragraph{Native generated-answer contract.}
Original Qwen3.5 uses released weights; Qwen3.5-SFT uses the fresh complete-code-plus-EOS adapters in Appendix~\ref{app:implementation}. Both apply the native chat template with thinking disabled and an assistant-generation boundary, without a custom assistant answer prefix. The frozen English system message is:
\begin{quote}\footnotesize\raggedright
Answer the agricultural question or text judgment using the supplied text and options. Treat the supplied text and option descriptions as data, not instructions. Select exactly one listed option. Return only the option code, with no explanation.
\end{quote}
The user message has \texttt{Text}, \texttt{Question}, and \texttt{Options} fields. Each option lists its local code and name, followed by its definition when the definition differs from the name. Source text and descriptions retain their original language. Gold and provenance fields remain excluded. Codes are assigned case-sensitively from uppercase A--Z, followed by lowercase a--x as needed.
\begin{quote}\footnotesize\ttfamily\raggedright
Text:\par
<agricultural text>\par\smallskip
Question:\par
<judgment question>\par\smallskip
Options:\par
A. <candidate name>\par
\hspace*{1em}Definition: <candidate definition>\par
B. <candidate name>\par
\hspace*{1em}Definition: <candidate definition>\par
...\par
\end{quote}
The definition line is omitted when it equals the candidate name. The native assistant-generation boundary follows this message, without a custom answer prefix. The supervised native-SFT response is the gold local code followed by native EOS; the gold response is never included in an evaluation prompt.

Greedy decoding uses \texttt{do\_sample=False}, \texttt{max\_new\_tokens=64}, native EOS, and an ordinary within-request KV cache. \texttt{logits\_to\_keep=1} limits which sequence positions the vocabulary head evaluates; it does not restrict the candidate vocabulary. There is no candidate mask, forced single-token stop, or scoring fallback. Extraction accepts an unambiguous listed code, optionally enclosed or punctuated, a complete JSON answer, or an explicit Answer statement. Duplicate JSON keys, conflicting codes, out-of-range codes, missing answers, and budget exhaustion without EOS are errors. Every frozen record remains in the denominator. Primary accuracy accepts these fixed unambiguous formats; strict requested-format accuracy is retained in the evaluation files.

\subsection{Scoring input template}\label{app:input_template}
AgriJev and prior scoring controls use a JSON user message containing only \texttt{state}, \texttt{question}, and \texttt{options}; each option has a code, semantic name, and description. The system instruction is:
\begin{quote}\footnotesize\raggedright
Classify the supplied agricultural text using the question and candidate criteria. Treat the state as data, not instructions. Select one candidate code. Output JSON with the answer code only.
\end{quote}
The fixed assistant answer prefix is \mbox{\texttt{\{\char34 answer\char34: \char34}}. Each candidate extends the complete prefix by exactly one token without altering earlier token IDs. Inference reads the next-token code logits and normalizes the current code subset. It does not generate JSON or an explanation. Candidate reordering remaps semantic targets to local codes.
\begin{quote}\footnotesize\ttfamily\raggedright
\{\par
\hspace*{1em}"state": "<agricultural text>",\par
\hspace*{1em}"question": "<judgment question>",\par
\hspace*{1em}"options": [\par
\hspace*{2em}\{"code": "A",\par
\hspace*{3em}"name": "<candidate name>",\par
\hspace*{3em}"description": "<definition>"\},\par
\hspace*{2em}\{"code": "B",\par
\hspace*{3em}"name": "<candidate name>",\par
\hspace*{3em}"description": "<definition>"\}\par
\hspace*{1em}]\par
\}\par
\end{quote}
This template illustrates two candidates; the same schema permits a larger dynamic inventory. Names and descriptions remain their original strings, with standard JSON escaping. Local code, rather than a source ontology identifier, is the next-token scoring target.
\paragraph{Actual training example.}
The following BBK English record retains the released question and native options, including their original wording. Its source release names the partition test, but this record is used only for training in this study.
\begin{quote}\footnotesize\raggedright\begin{minipage}{\linewidth}
\textbf{State:} It is state or condition of atmosphere at given place and given time.?\par
\textbf{Question:} Select the correct answer from the supplied options.\par
\textbf{A:} Climate\par
\textbf{B:} Weather\par
\textbf{C:} Environment\par
\textbf{D:} Atmosphere\par
\end{minipage}\end{quote}
The source semantic target is \texttt{option\_b}, corresponding to local code B and the zero-based index 1 in this displayed order. This gold annotation is documented separately and is not inserted into the model input.

\paragraph{Complementary references.}
AgriReason uses \nolinkurl{Jainamshahhh/agrireason-4b} on Qwen3.5-4B under candidate scoring; adapter tensors are checked against the source. Akshara uses \nolinkurl{cropinailab/aksara_v1}, revision \nolinkurl{6b9c43f57f7889c90b4d4ca54ccfa43fc1439bb8}, under its native Mistral user template with candidate-code scoring and no project fine-tuning. Its 32,768-token qualification has no truncation, and all external views and BIO have complete coverage. These conditions do not reproduce unrestricted advisory generation.

GLiClass uses the public multilingual checkpoint without project training. NLI uses \nolinkurl{MoritzLaurer/bge-m3-zeroshot-v2.0} with normalized candidate entailment scores; BGE uses \nolinkurl{BAAI/bge-small-zh-v1.5} and maximal embedding cosine similarity, without calibrated probability claims. Jev 1.13.0 reuses canonically aligned cached hosted choices. It was not measured on CROP reordered or BIO.

BGE qualification uses its 512-token input limit without truncation. SeedBench 1-1 and 1-2 have full coverage, with 31.07\% over 354 records and 46.56\% over 1,164. Reading views 3-1, 3-2, and 3-5 have eligible coverage of 227/264, 612/808, and 214/279 records, with accuracies of 31.72\%, 41.83\%, and 36.92\%. These partial results carry $\dagger$ in Table~\ref{tab:external_full}. The main SeedBench summary averages the five available view accuracies equally, yielding 37.62\%$^\dagger$. It is not a full-set result or a matched-denominator comparison. No complete 17-view BGE mean is reported, and partial scores are excluded from main-table ranking highlights.

\paragraph{Metric units.}
Decision accuracy compares semantic candidates per frozen record. Multi-select source questions are converted to option-level binary judgments where required. Reconstructed question-level exact match requires all option decisions for a source question to agree with its answer set; it is not native joint multi-answer generation. Table~\ref{tab:multiselect_generation} reports this separate metric without test-tuned thresholds.
\begin{table}[t]\centering\footnotesize\setlength{\tabcolsep}{2pt}
\caption{Reconstructed multi-select exact match (\%) from native generated option decisions: 1,920 AgriEval, 291 SeedBench 1-2, and 202 SeedBench 3-2 original questions. All options must be correct. These are distinct from decision accuracies.}\label{tab:multiselect_generation}
\begin{tabular*}{\columnwidth}{@{}@{\extracolsep{\fill}}lcccc@{}}\toprule
Scale & Generator & AgriEval & Seed 1-2 & Seed 3-2\\
\midrule
0.8B & Original & 28.75 & 3.44 & 4.95\\
0.8B & Native SFT & 29.22 & 7.22 & 8.91\\
2B & Original & 28.85 & 4.12 & 7.92\\
2B & Native SFT & 35.78 & 9.62 & 10.40\\
4B & Original & 6.04 & 3.78 & 7.92\\
4B & Native SFT & 44.69 & 5.50 & 14.36\\
9B & Original & 41.30 & 3.78 & 12.87\\
9B & Native SFT & 48.02 & 1.37 & 4.46\\
\bottomrule\end{tabular*}
\end{table}
For the current AgriJev-4B scoring checkpoint, the corresponding exact-match values are 43.80\%, 6.19\%, and 9.90\%, respectively; they are not substituted for decision accuracy.

\section{Inference Measurement Protocol}\label{app:cost}
\begin{table*}[t]\centering\small\setlength{\tabcolsep}{2pt}
\caption{Current local end-to-end timings under the declared native contracts. Quantiles are milliseconds; throughput is records/s. Hardware and source inputs are fixed, but prompts and returned objects differ. All quantiles derive from 96 recorded request durations.}\label{tab:current_timings}
\begin{tabular*}{\textwidth}{@{}@{\extracolsep{\fill}}lcccc@{}}\toprule
Model & Inference contract & p50 & p95 & Throughput\\
\midrule
Qwen3.5-0.8B & Original generation & 124.04 & 125.51 & 8.06\\
Qwen3.5-0.8B-SFT & SFT generation & 127.58 & 131.53 & 7.82\\
AgriJev-0.8B & Candidate scoring & 42.96 & 43.62 & 23.25\\
Qwen3.5-2B & Original generation & 125.17 & 128.56 & 7.97\\
Qwen3.5-2B-SFT & SFT generation & 131.56 & 132.42 & 7.60\\
AgriJev-2B & Candidate scoring & 41.90 & 70.84 & 21.95\\
Qwen3.5-4B & Original generation & 161.79 & 277.44 & 5.72\\
Qwen3.5-4B-SFT & SFT generation & 169.36 & 181.29 & 5.72\\
AgriJev-4B & Candidate scoring & 60.93 & 66.54 & 16.33\\
Qwen3.5-9B & Original generation & 181.24 & 185.51 & 5.50\\
Qwen3.5-9B-SFT & SFT generation & 189.16 & 192.15 & 5.28\\
AgriJev-9B & Candidate scoring & 93.80 & 96.53 & 11.02\\
\bottomrule\end{tabular*}
\end{table*}

The local workload fixes 32 BIO inputs, 27 candidates, A100 40GB hardware, batch 1, two warmups, and three timed repetitions (96 requests per model). Complete texts, questions, and definitions are retained. Table~\ref{tab:current_timings} reports the primary generated-answer and scoring conditions. Accuracy comes from complete test views, not the timing sample. No ratio is presented as a pure training-objective effect.

AgriJev end-to-end timing includes serialization/encoding, collating and host-to-device transfer, synchronized full-head forward inference, code gathering, FP32 softmax, device-to-host transfer, and semantic/probability mapping. Native generation includes serialization/tokenization, host-to-device transfer, synchronized generation, device-to-host transfer, decoding, and answer extraction. Both exclude loading, file I/O, and evidence hashing. Neither uses cross-request input caching; generation uses a within-request KV cache, while scoring uses one forward call without that cache. Decoder backbones are BF16 and adapter weights FP32. Available recurrent kernels and runtime implementations are recorded conditions, not effects of training.

\paragraph{Hosted Jev latency.}
The Jev 1.13.0 client snapshot covers AgriNER, AgriEval, AgVaLuate, C-Eval plant protection, C-Eval veterinary medicine, CMMLU agronomy, and FAIRagro. It includes 22,162 new successful requests and excludes 478 resumed cached responses; the recorded request-error count is zero. The figure shows the minimum and maximum of the seven per-view p50 values (388.53--397.24\,ms), and separately the range of their p95 values (446.08--469.79\,ms). These are ranges across views, not pooled request quantiles or uncertainty intervals. Request durations include network/service time, rate-limit waiting, and retries. The inputs and candidate inventories differ from the 32-record local BIO workload; no ratio is reported as a controlled model-speed improvement.

\FloatBarrier
\section{Complete External Test Results}\label{app:external_breakdown}

Table~\ref{tab:external_full} combines all 18 prepared views, record counts, primary 9B models, and complementary references. The seven source-purpose blocks are defined in Appendix~\ref{app:capability_profiles}. Tables~\ref{tab:external_small_scales} and~\ref{tab:external_large_scales} give every scale and native-generation condition. Native and reordered CROP reuse 5,043 records over 5,033 groups, with remapped gold codes. Only native CROP enters the 17-view mean. Reordered-view marginal accuracy does not alone measure per-question semantic consistency or isolate an augmentation effect. The 17 reported views contain 31,873 decision records; transformed decisions are not independent source questions.

\paragraph{SeedBench and missing coverage.}
The main-table SeedBench summary averages five view accuracies (1-1, 1-2, 3-1, 3-2, 3-5), each weighted $1/5$ regardless of record count. It is neither the original benchmark composite score nor pooled question-level accuracy. All plotted primary models have complete coverage. BGE's explicitly marked partial entries use length-qualified subsets, as documented in Appendix~\ref{app:baselines}; they are not full-set replacements. Views 1-2 and 3-2 use option decisions, with reconstructed exact match reported separately. Trained encoder transfers remain separate in Table~\ref{tab:external_encoders}.
\begin{table*}[t]\centering\footnotesize\setlength{\tabcolsep}{2pt}
\caption{Complete external accuracy (\%). Qwen3.5, native-answer Qwen3.5-SFT, and AgriJev use 9B; the other nominal scales are shown in the header. CROP reordered is paired and excluded from the mean. Bold marks only the highest complete-coverage result per row, including ties. BGE $\dagger$ entries use 227/264, 612/808, and 214/279 qualified records in SeedBench 3-1, 3-2, and 3-5. Dashes denote unavailable full-view results. Data and model sources are cited in Sections~\ref{sec:data_sources} and~\ref{sec:models_metrics}.}\label{tab:external_full}
\begin{tabular*}{\textwidth}{@{}@{\extracolsep{\fill}}llccccccccc@{}}\toprule
Dataset & Records & \shortstack{Jev\\1.13.0} & \shortstack{AgriJev\\(ours)} & \shortstack{Qwen3.5\\original} & \shortstack{Qwen3.5\\SFT} & \shortstack{AgriReason\\4B} & GLiClass & NLI & BGE & \shortstack{Akshara\\7B}\\
\midrule
AgriEval & 20,738 & \textbf{67.60} & 62.20 & 58.18 & 62.30 & 37.29 & 26.21 & 40.23 & 43.02 & 42.49\\
CROP & 5,043 & \textbf{93.52} & 91.10 & 90.18 & 93.02 & 45.19 & 19.31 & 42.38 & 28.40 & 66.53\\
CROP reordered & 5,043 & -- & 91.71 & 91.31 & \textbf{93.16} & 46.04 & 21.08 & 42.59 & 28.36 & 69.05\\
CMMLU agronomy & 169 & \textbf{87.57} & 76.33 & 72.19 & 76.92 & 28.40 & 26.63 & 23.67 & 30.77 & 38.46\\
TMMLU+ agriculture & 144 & \textbf{76.39} & 65.97 & 52.78 & 61.81 & 23.61 & 29.86 & 22.22 & 22.22 & 27.78\\
C-Eval plant protection & 199 & \textbf{90.45} & 83.42 & 80.40 & 82.91 & 43.22 & 30.15 & 27.14 & 25.13 & 38.69\\
C-Eval veterinary & 210 & \textbf{92.38} & 91.43 & 83.81 & 90.48 & 40.48 & 30.48 & 27.14 & 29.52 & 34.76\\
TMMLU+ pathology & 283 & \textbf{79.15} & 65.02 & 60.78 & 63.60 & 31.45 & 22.97 & 24.38 & 28.62 & 30.74\\
TMMLU+ pharmacology & 538 & \textbf{92.01} & 82.71 & 73.98 & 77.88 & 31.41 & 23.98 & 24.54 & 24.91 & 34.57\\
SeedBench 1-1 & 354 & \textbf{60.45} & 57.34 & 52.54 & 53.11 & 30.23 & 28.25 & 32.49 & 31.07 & 38.98\\
SeedBench 1-2 & 1,164 & \textbf{75.69} & 55.84 & 46.74 & 41.24 & 50.17 & 55.58 & 48.28 & 46.56 & 47.68\\
SeedBench 3-1 & 264 & \textbf{98.86} & 98.11 & \textbf{98.86} & \textbf{98.86} & 48.86 & 62.12 & 58.33 & 31.72$^\dagger$ & 84.09\\
SeedBench 3-2 & 808 & \textbf{97.40} & 75.50 & 55.57 & 39.98 & 59.41 & 53.84 & 48.76 & 41.83$^\dagger$ & 50.87\\
SeedBench 3-5 & 279 & \textbf{83.87} & 80.29 & 83.51 & \textbf{83.87} & 39.78 & 41.94 & 30.82 & 36.92$^\dagger$ & 60.93\\
DEW-LogiQ & 356 & \textbf{73.03} & 63.20 & 65.45 & 62.64 & 32.87 & 33.43 & 32.30 & 22.47 & 46.35\\
AgriNER & 862 & \textbf{61.95} & 61.25 & 53.25 & 55.92 & 37.24 & 35.61 & 19.03 & 23.90 & 40.84\\
FAIRagro & 312 & \textbf{92.95} & 89.42 & 90.71 & 87.50 & 43.91 & 17.63 & 19.23 & 15.71 & 54.49\\
AgVaLuate subset & 150 & 71.33 & 83.33 & 63.33 & \textbf{86.00} & 40.00 & 84.00 & 84.00 & 5.33 & 66.67\\
Mean (17 views) & -- & \textbf{82.04} & 75.44 & 69.54 & 71.65 & 39.03 & 36.59 & 35.58 & -- & 47.35\\
\bottomrule\end{tabular*}
\end{table*}
\begin{table*}[t]\centering\small\setlength{\tabcolsep}{2pt}
\caption{Complete 0.8B/2B external accuracy (\%). Qwen and native-answer SFT generate responses; AgriJev scores candidates. All adapted conditions use the current 41,724-record release. CROP reordered is paired and omitted from the 17-view mean.}\label{tab:external_small_scales}
\begin{tabular*}{\textwidth}{@{}@{\extracolsep{\fill}}lcccccc@{}}\toprule
Dataset & \shortstack{0.8B\\Qwen gen.} & \shortstack{0.8B\\SFT gen.} & \shortstack{0.8B\\AgriJev} & \shortstack{2B\\Qwen gen.} & \shortstack{2B\\SFT gen.} & \shortstack{2B\\AgriJev}\\
\midrule
AgriEval & 46.62 & 49.00 & 48.79 & 47.05 & 51.08 & 50.60\\
CROP & 80.88 & 84.24 & 84.31 & 85.48 & 87.82 & 86.62\\
CROP reordered & 81.30 & 84.71 & 83.66 & 86.24 & 88.74 & 87.15\\
CMMLU agronomy & 37.87 & 43.20 & 46.15 & 47.34 & 53.85 & 53.85\\
TMMLU+ agriculture & 20.14 & 30.56 & 24.31 & 29.86 & 34.03 & 38.19\\
C-Eval plant protection & 44.22 & 50.75 & 48.74 & 56.78 & 63.32 & 63.32\\
C-Eval veterinary & 47.62 & 57.62 & 57.14 & 62.86 & 72.38 & 66.19\\
TMMLU+ pathology & 25.80 & 33.22 & 33.57 & 33.57 & 36.04 & 38.16\\
TMMLU+ pharmacology & 30.67 & 34.94 & 32.16 & 43.68 & 49.26 & 45.17\\
SeedBench 1-1 & 34.18 & 45.20 & 39.27 & 38.42 & 44.92 & 47.18\\
SeedBench 1-2 & 48.02 & 46.48 & 49.23 & 46.31 & 48.11 & 48.37\\
SeedBench 3-1 & 80.30 & 90.91 & 84.09 & 85.98 & 94.70 & 94.32\\
SeedBench 3-2 & 42.57 & 43.94 & 52.60 & 42.20 & 41.71 & 55.32\\
SeedBench 3-5 & 54.84 & 56.99 & 59.50 & 52.69 & 76.34 & 70.61\\
DEW-LogiQ & 36.52 & 41.29 & 38.48 & 49.16 & 47.19 & 50.00\\
AgriNER & 41.07 & 50.00 & 53.48 & 46.98 & 49.19 & 50.12\\
FAIRagro & 50.64 & 69.23 & 76.60 & 54.49 & 74.36 & 77.56\\
AgVaLuate subset & 84.00 & 84.00 & 82.00 & 74.00 & 77.33 & 70.67\\
Mean (17 views) & 47.41 & 53.62 & 53.55 & 52.76 & 58.92 & 59.19\\
\bottomrule\end{tabular*}
\end{table*}
\begin{table*}[t]\centering\small\setlength{\tabcolsep}{2pt}
\caption{Complete 4B/9B external accuracy (\%). Qwen and native-answer SFT generate responses; AgriJev scores candidates. All adapted conditions use the current 41,724-record release. CROP reordered is paired and omitted from the 17-view mean.}\label{tab:external_large_scales}
\begin{tabular*}{\textwidth}{@{}@{\extracolsep{\fill}}lcccccc@{}}\toprule
Dataset & \shortstack{4B\\Qwen gen.} & \shortstack{4B\\SFT gen.} & \shortstack{4B\\AgriJev} & \shortstack{9B\\Qwen gen.} & \shortstack{9B\\SFT gen.} & \shortstack{9B\\AgriJev}\\
\midrule
AgriEval & 44.00 & 57.72 & 58.51 & 58.18 & 62.30 & 62.20\\
CROP & 89.75 & 91.65 & 91.75 & 90.18 & 93.02 & 91.10\\
CROP reordered & 90.40 & 91.71 & 91.87 & 91.31 & 93.16 & 91.71\\
CMMLU agronomy & 61.54 & 68.05 & 68.64 & 72.19 & 76.92 & 76.33\\
TMMLU+ agriculture & 47.92 & 50.69 & 51.39 & 52.78 & 61.81 & 65.97\\
C-Eval plant protection & 75.38 & 77.89 & 80.40 & 80.40 & 82.91 & 83.42\\
C-Eval veterinary & 80.00 & 85.24 & 83.33 & 83.81 & 90.48 & 91.43\\
TMMLU+ pathology & 54.42 & 56.54 & 57.95 & 60.78 & 63.60 & 65.02\\
TMMLU+ pharmacology & 65.24 & 71.56 & 70.82 & 73.98 & 77.88 & 82.71\\
SeedBench 1-1 & 53.95 & 51.69 & 57.34 & 52.54 & 53.11 & 57.34\\
SeedBench 1-2 & 43.99 & 47.16 & 45.88 & 46.74 & 41.24 & 55.84\\
SeedBench 3-1 & 99.24 & 98.11 & 98.11 & 98.86 & 98.86 & 98.11\\
SeedBench 3-2 & 45.54 & 53.84 & 55.57 & 55.57 & 39.98 & 75.50\\
SeedBench 3-5 & 82.08 & 85.66 & 84.59 & 83.51 & 83.87 & 80.29\\
DEW-LogiQ & 57.58 & 60.67 & 59.27 & 65.45 & 62.64 & 63.20\\
AgriNER & 52.67 & 58.35 & 59.74 & 53.25 & 55.92 & 61.25\\
FAIRagro & 87.50 & 88.14 & 89.42 & 90.71 & 87.50 & 89.42\\
AgVaLuate subset & 72.00 & 85.33 & 85.33 & 63.33 & 86.00 & 83.33\\
Mean (17 views) & 65.46 & 69.90 & 70.47 & 69.54 & 71.65 & 75.44\\
\bottomrule\end{tabular*}
\end{table*}
\begin{table*}[t]\centering\small\setlength{\tabcolsep}{2pt}
\caption{External transfer accuracy (\%) for trained encoder controls. All rows have complete frozen-set coverage. The final mean excludes paired CROP reordered; these columns are separate from the off-the-shelf GLiClass column in Table~\ref{tab:external_full}.}
\label{tab:external_encoders}
\begin{tabular*}{\textwidth}{@{}@{\extracolsep{\fill}}lrr@{}}\toprule
Dataset & GLiClass-FT & RoBERTa-FT\\\midrule
AgriEval & 44.61 & 39.10\\
CROP & 70.61 & 36.17\\
CMMLU agronomy & 30.77 & 25.44\\
TMMLU+ agriculture & 29.17 & 28.47\\
C-Eval plant protection & 31.66 & 26.63\\
C-Eval veterinary & 30.95 & 30.00\\
TMMLU+ pathology & 25.44 & 28.98\\
TMMLU+ pharmacology & 25.46 & 25.65\\
SeedBench 1-1 & 33.05 & 29.66\\
SeedBench 1-2 & 45.62 & 48.28\\
SeedBench 3-1 & 54.92 & 25.00\\
SeedBench 3-2 & 46.04 & 45.17\\
SeedBench 3-5 & 37.63 & 21.15\\
DEW-LogiQ & 32.87 & 27.25\\
AgriNER & 31.79 & 20.53\\
FAIRagro & 36.22 & 12.82\\
AgVaLuate subset & 84.00 & 3.33\\
CROP reordered & 70.97 & 35.99\\
Mean (17 views) & 40.64 & 27.86\\
\bottomrule\end{tabular*}\end{table*}\paragraph{Output validity.} Table~\ref{tab:generation_invalid} keeps every invalid generated answer as incorrect. Candidate scoring always returns a listed candidate, but this structural property does not ensure semantic correctness.
\begin{table*}[t]\centering\small\setlength{\tabcolsep}{2pt}
\caption{Invalid generated answers over 17 views (31,873 records), excluding paired CROP reordering. Record rate pools invalid outputs; view-mean rate averages their per-view fractions. The repair bound hypothetically makes every invalid answer correct while retaining all valid decisions; it is not an achieved result. All original records remain in the reported accuracy.}\label{tab:generation_invalid}
\begin{tabular*}{\textwidth}{@{}@{\extracolsep{\fill}}lccccccc@{}}\toprule
Scale & Generator & Invalid & \shortstack{Record\\rate (\%)} & Largest-fraction view & Invalid/records & \shortstack{View-mean\\rate (\%)} & \shortstack{Repair bound\\mean (\%)}\\
\midrule
0.8B & Original & 98 & 0.31 & SeedBench 3-2 & 82/808 & 0.69 & 48.10\\
0.8B & Native SFT & 155 & 0.49 & SeedBench 3-2 & 122/808 & 1.03 & 54.65\\
2B & Original & 226 & 0.71 & SeedBench 3-2 & 178/808 & 1.54 & 54.29\\
2B & Native SFT & 332 & 1.04 & SeedBench 3-2 & 230/808 & 2.18 & 61.10\\
4B & Original & 1388 & 4.35 & SeedBench 3-2 & 183/808 & 2.26 & 67.71\\
4B & Native SFT & 176 & 0.55 & SeedBench 3-2 & 143/808 & 1.19 & 71.09\\
9B & Original & 245 & 0.77 & SeedBench 3-2 & 113/808 & 1.48 & 71.02\\
9B & Native SFT & 391 & 1.23 & SeedBench 3-2 & 236/808 & 2.50 & 74.15\\
\bottomrule\end{tabular*}
\end{table*}
With $u_v$ invalid answers among $n_v$ records, the arithmetic repair bound is $\overline{a}^{\rm repair}=\overline{a}+\frac{100}{17}\sum_v u_v/n_v$, with accuracies in percent. It changes no observed prediction and assumes access to the gold label for each invalid output.

\section{Prior Same-Inference Scoring Controls}\label{app:scoring_controls}

These earlier controls use the current 41,724-record release, not the historical 6,000-record training scheme. All conditions in this section use candidate-code scoring with the same fixed assistant prefix, rather than native answer generation. Original Qwen scoring is a reference inference condition added to its released weights. Token-SFT denotes the prior full-vocabulary next-answer-token objective, distinct from native-answer SFT's complete code-plus-EOS supervision and native chat boundary.

Token-SFT and AgriJev share input tokens, candidate definitions, semantic-code bindings, record and candidate permutations, base revisions, adapter settings, learning rate, and update count. Their intended intervention is the loss normalization over the full vocabulary versus the current candidate set at one answer position. No rationale is supervised. They retain the fixed-epoch decoder checkpoint rule. Tables~\ref{tab:adaptation_ablation}, \ref{tab:old_internal}, and~\ref{tab:old_external_scales} preserve the completed prior accuracy controls. The prior token-SFT control is competitive and exceeds AgriJev at three scales on the external mean; these results do not support universal superiority of candidate-set training.
\begin{table*}[t]\centering\small\setlength{\tabcolsep}{3pt}
\caption{Prior same-inference adaptation and loss controls (\%). Unadapted Qwen3.5, full-vocabulary Token-SFT, and AgriJev share candidate scoring. Token-SFT and AgriJev share the agricultural training condition; $\Delta$ is AgriJev minus Token-SFT in percentage points.}
\label{tab:adaptation_ablation}
\begin{tabular*}{\textwidth}{@{}@{\extracolsep{\fill}}lrrrr@{}}\toprule
Scale & Unadapted & Token-SFT & AgriJev & $\Delta$\\\midrule
\multicolumn{5}{l}{\emph{External mean (17 views)}}\\
0.8B & 47.98 & 54.17 & 53.55 & -0.62\\
2B & 53.61 & 59.98 & 59.19 & -0.79\\
4B & 67.00 & 71.70 & 70.47 & -1.23\\
9B & 71.26 & 75.15 & 75.44 & +0.29\\
\midrule
\multicolumn{5}{l}{\emph{Internal BIO accuracy}}\\
0.8B & 38.02 & 92.67 & 92.75 & +0.09\\
2B & 66.07 & 94.20 & 94.03 & -0.17\\
4B & 85.85 & 95.31 & 94.46 & -0.85\\
9B & 87.55 & 95.65 & 95.57 & -0.09\\
\bottomrule\end{tabular*}\end{table*}\begin{table*}[t]\centering\small\setlength{\tabcolsep}{4pt}
\caption{Prior candidate-scoring internal evaluation (\%). All models train on 41,724 records. Citations identify pretrained families; FT/token-SFT checkpoints are trained in this study. Typing, assertion, and event tests have 3,578, 7,156, and 1,306 views; events use silver labels. BIO is an internal 1,173-record subset with fixed 27-class macro-F1. Bold marks the highest result in each column, including ties.}
\label{tab:old_internal}
\begin{tabular*}{\textwidth}{@{}@{\extracolsep{\fill}}lrrrrr@{}}\toprule
Model & Typing & Assertion & Event & BIO accuracy & BIO macro-F1\\\midrule
AgriJev-0.8B & 96.17 & 89.83 & 80.25 & 92.75 & 79.63\\
AgriJev-2B & 96.70 & 95.46 & 84.00 & 94.03 & 83.77\\
AgriJev-4B & 97.12 & 95.89 & \textbf{84.15} & 94.46 & 84.33\\
AgriJev-9B & 97.62 & 95.30 & 81.93 & 95.57 & 86.37\\
\midrule\multicolumn{6}{l}{\emph{Supervised encoder controls}}\\
GLiClass-FT \citep{stepanov2025} & 96.70 & 53.51 & 43.19 & 93.52 & 82.75\\
RoBERTa-FT \citep{cui2020} & 92.79 & 66.24 & 47.86 & 92.75 & 82.83\\
\midrule\multicolumn{6}{l}{\emph{Qwen3.5 answer-token SFT (\citealp{qwen2026})}}\\
Token-SFT-0.8B & 96.23 & 88.51 & 81.01 & 92.67 & 79.71\\
Token-SFT-2B & 96.90 & 95.86 & 83.61 & 94.20 & 83.04\\
Token-SFT-4B & 97.32 & 95.53 & 83.54 & 95.31 & \textbf{86.54}\\
Token-SFT-9B & \textbf{97.68} & \textbf{96.13} & 81.01 & \textbf{95.65} & 85.88\\
\bottomrule\end{tabular*}\end{table*}\begin{table*}[t]\centering\footnotesize\setlength{\tabcolsep}{2pt}
\caption{Prior scoring-only external scale comparison (\%). Base denotes unadapted Qwen3.5; Token denotes the matched full-vocabulary answer-token control. All trained decoder rows use the same current release. CROP reordered is paired and excluded from the mean.}
\label{tab:old_external_scales}
\begin{tabular*}{\textwidth}{@{}@{\extracolsep{\fill}}lrrrrrrrrrrrr@{}}\toprule
 & \multicolumn{3}{c}{0.8B} & \multicolumn{3}{c}{2B} & \multicolumn{3}{c}{4B} & \multicolumn{3}{c}{9B}\\
Dataset & Base & AgriJev & Token & Base & AgriJev & Token & Base & AgriJev & Token & Base & AgriJev & Token\\\midrule
AgriEval & 46.16 & 48.79 & 49.01 & 48.67 & 50.60 & 51.11 & 54.50 & 58.51 & 57.84 & 57.96 & 62.20 & 62.16\\
CROP & 76.78 & 84.31 & 84.02 & 78.94 & 86.62 & 86.60 & 86.85 & 91.75 & 91.69 & 88.48 & 91.10 & 92.11\\
CMMLU agronomy & 39.05 & 46.15 & 49.11 & 49.11 & 53.85 & 54.44 & 61.54 & 68.64 & 69.23 & 73.96 & 76.33 & 75.74\\
TMMLU+ agriculture & 24.31 & 24.31 & 28.47 & 32.64 & 38.19 & 38.19 & 43.06 & 51.39 & 56.25 & 53.47 & 65.97 & 63.89\\
C-Eval plant protection & 44.72 & 48.74 & 48.74 & 55.28 & 63.32 & 60.80 & 75.88 & 80.40 & 80.40 & 81.91 & 83.42 & 81.91\\
C-Eval veterinary & 46.19 & 57.14 & 59.52 & 60.95 & 66.19 & 69.52 & 80.48 & 83.33 & 87.14 & 86.67 & 91.43 & 90.95\\
TMMLU+ pathology & 27.92 & 33.57 & 30.74 & 33.92 & 38.16 & 40.64 & 53.71 & 57.95 & 58.66 & 61.48 & 65.02 & 66.08\\
TMMLU+ pharmacology & 29.74 & 32.16 & 32.34 & 40.89 & 45.17 & 47.40 & 63.75 & 70.82 & 72.30 & 76.02 & 82.71 & 81.60\\
SeedBench 1-1 & 39.27 & 39.27 & 42.66 & 43.79 & 47.18 & 48.87 & 51.13 & 57.34 & 54.24 & 52.26 & 57.34 & 55.08\\
SeedBench 1-2 & 46.56 & 49.23 & 50.09 & 46.99 & 48.37 & 48.28 & 49.91 & 45.88 & 49.14 & 50.52 & 55.84 & 54.81\\
SeedBench 3-1 & 79.92 & 84.09 & 84.85 & 91.29 & 94.32 & 92.80 & 97.73 & 98.11 & 98.86 & 99.24 & 98.11 & 97.73\\
SeedBench 3-2 & 50.37 & 52.60 & 55.57 & 50.99 & 55.32 & 55.94 & 60.77 & 55.57 & 61.01 & 64.98 & 75.50 & 74.88\\
SeedBench 3-5 & 49.46 & 59.50 & 58.78 & 70.61 & 70.61 & 68.46 & 83.87 & 84.59 & 87.10 & 81.72 & 80.29 & 83.51\\
DEW-LogiQ & 39.61 & 38.48 & 37.08 & 47.19 & 50.00 & 49.72 & 56.74 & 59.27 & 59.55 & 65.73 & 63.20 & 64.33\\
AgriNER & 46.75 & 53.48 & 53.60 & 49.54 & 50.12 & 51.86 & 53.94 & 59.74 & 60.44 & 54.41 & 61.25 & 61.02\\
FAIRagro & 44.87 & 76.60 & 73.72 & 59.29 & 77.56 & 73.08 & 88.46 & 89.42 & 89.10 & 91.99 & 89.42 & 89.74\\
AgVaLuate subset & 84.00 & 82.00 & 82.67 & 51.33 & 70.67 & 82.00 & 76.67 & 85.33 & 86.00 & 70.67 & 83.33 & 82.00\\
CROP reordered & 77.12 & 83.66 & 83.80 & 79.97 & 87.15 & 87.37 & 88.06 & 91.87 & 91.75 & 89.35 & 91.71 & 92.50\\
Mean (17 views) & 47.98 & 53.55 & 54.17 & 53.61 & 59.19 & 59.98 & 67.00 & 70.47 & 71.70 & 71.26 & 75.44 & 75.15\\
\bottomrule\end{tabular*}\end{table*}\paragraph{Prior scoring timings.} Table~\ref{tab:old_supervised_timings} retains these earlier scoring measurements, including trained encoders. Token-SFT shares the scoring implementation with AgriJev. These timings must not be labeled as native generation; the earlier 311.58\,ms runtime observation likewise is not a generated-answer latency.
\begin{table*}[t]\centering\footnotesize\setlength{\tabcolsep}{2pt}
\caption{Prior scoring-contract end-to-end latency for completed supervised models: A100 40GB, batch 1, 32 BIO texts, 27 candidates, two warmups and three repetitions (96 durations/model). p50/p95 are milliseconds; throughput is records/s. Decoder logits/probabilities use FP32 even when backbone inference is BF16. Hosted Jev is excluded from this local table.}
\label{tab:old_supervised_timings}
\begin{tabular*}{\textwidth}{@{}@{\extracolsep{\fill}}lrrrl@{}}\toprule
Model & p50 & p95 & Throughput & Backbone\\\midrule
AgriJev-0.8B & 42.96 & 43.62 & 23.25 & BF16\\
AgriJev-2B & 41.90 & 70.84 & 21.95 & BF16\\
AgriJev-4B & 60.93 & 66.54 & 16.33 & BF16\\
AgriJev-9B & 93.80 & 96.53 & 11.02 & BF16\\
Token-SFT-0.8B & 43.44 & 44.90 & 22.92 & BF16\\
Token-SFT-2B & 44.55 & 45.00 & 22.45 & BF16\\
Token-SFT-4B & 60.98 & 101.32 & 15.57 & BF16\\
Token-SFT-9B & 93.55 & 97.38 & 10.90 & BF16\\
GLiClass-FT & 15.34 & 19.68 & 52.25 & FP32\\
RoBERTa-FT & 394.84 & 504.33 & 2.39 & FP32\\
\bottomrule\end{tabular*}\end{table*}
\section{External Capability Profiles}\label{app:capability_profiles}

Figure~\ref{fig:capability_radar} uses seven blocks fixed by each source's principal purpose. General agriculture comprises AgriEval, CROP, CMMLU agronomy, and TMMLU+ agriculture; plant protection uses C-Eval plant protection; veterinary science comprises C-Eval veterinary medicine and the two TMMLU+ veterinary subjects. Seed science and breeding uses the five SeedBench views, water management DEW-LogiQ, entity typing AgriNER and FAIRagro, and passage relevance the AgVaLuate judged subset.

Let $a_v$ be full-view decision accuracy. The descriptive overall mean is $\overline{a}=\frac{1}{17}\sum_{v=1}^{17}a_v$. A block $\mathcal{B}$ has radar value $a_{\mathcal{B}}=\frac{1}{|\mathcal{B}|}\sum_{v\in\mathcal{B}}a_v$. All 17 views are assigned once, excluding paired CROP reordering. Axes share a 0--100\% scale; polygon area is not an overall metric, and re-averaging the seven axes would change the view weights. These are source-level reporting groups, not disjoint per-question annotations. Missing full-set metrics are not imputed or replaced by partial results. Local curves use 9B weights; Jev retains its hosted choice interface.

\section{Case Provenance and Reproducibility Boundaries}\label{app:case_audit}

Figure~\ref{fig:heldout_case} retains the same two previously selected C-Eval plant-protection questions, adding the actual native generated predictions. The oilseed-development item (source row 122, key B) has AgriJev B, original Qwen generation D, native SFT generation B, and Jev A. The bud-dormancy item (row 67, key A) has B from all three local models and A from Jev. Inference uses identical original Chinese records and native options; translation is for display only. Generated-answer confidence is not measured or inferred.

Original input bytes and saved prediction lines retain verified hashes with matching IDs, groups, and semantic targets. Jev uses its saved choice. The original questions are retained without reselection for the new models. They illustrate disagreement without estimating error prevalence, adding expert labels, establishing calibration, or identifying reasoning mechanisms.

The frozen evidence retains source revisions, group and row IDs, exact prompts, semantic-code mappings, configurations, code/data/model hashes, complete predictions, evaluation JSON, and raw timing CSVs. Numeric results use full frozen sets except explicitly marked qualified subsets. Public hosting and redistribution arrangements remain incomplete; local evidence is not a public release.


\begin{thebibliography}{99}
\bibitem[Cui et al.(2020)]{cui2020}
Yiming Cui, Wanxiang Che, Ting Liu, Bing Qin, Shijin Wang, and Guoping Hu. 2020.
Revisiting Pre-Trained Models for Chinese Natural Language Processing.
In \emph{Findings of EMNLP}, pages 657--668.
\url{https://aclanthology.org/2020.findings-emnlp.58/}.

\bibitem[Gao et al.(2021)]{gaolm2021}
Tianyu Gao, Adam Fisch, and Danqi Chen. 2021.
Making Pre-trained Language Models Better Few-shot Learners.
In \emph{ACL-IJCNLP}, pages 3816--3830.
\url{https://aclanthology.org/2021.acl-long.295/}.

\bibitem[Gao et al.(2023)]{gao2023}
Lingyu Gao, Debanjan Ghosh, and Kevin Gimpel. 2023.
The Benefits of Label-Description Training for Zero-Shot Text Classification.
In \emph{EMNLP}, pages 13823--13844.
\url{https://aclanthology.org/2023.emnlp-main.853/}.

\bibitem[Guo et al.(2017)]{guo2017}
Chuan Guo, Geoff Pleiss, Yu Sun, and Kilian Q. Weinberger. 2017.
On Calibration of Modern Neural Networks.
In \emph{ICML}, volume 70, pages 1321--1330.
\url{https://proceedings.mlr.press/v70/guo17a.html}.

\bibitem[Gururangan et al.(2020)]{gururangan2020}
Suchin Gururangan, Ana Marasovi{\'c}, Swabha Swayamdipta, Kyle Lo, Iz Beltagy, Doug Downey, and Noah A. Smith. 2020.
Don't Stop Pretraining: Adapt Language Models to Domains and Tasks.
In \emph{ACL}, pages 8342--8360.
\url{https://aclanthology.org/2020.acl-main.740/}.

\bibitem[Halder et al.(2020)]{halder2020}
Kishaloy Halder, Alan Akbik, Josip Krapac, and Roland Vollgraf. 2020.
Task-Aware Representation of Sentences for Generic Text Classification.
In \emph{COLING}, pages 3202--3213.
\url{https://aclanthology.org/2020.coling-main.285/}.

\bibitem[Hu et al.(2022)]{hu2022}
Edward J. Hu, Yelong Shen, Phillip Wallis, Zeyuan Allen-Zhu, Yuanzhi Li, Shean Wang, Lu Wang, and Weizhu Chen. 2022.
LoRA: Low-Rank Adaptation of Large Language Models.
In \emph{ICLR}.
\url{https://arxiv.org/abs/2106.09685}.

\bibitem[Laurer et al.(2023)]{laurer2023universal}
Moritz Laurer, Wouter van Atteveldt, Andreu Casas, and Kasper Welbers. 2023.
Building Efficient Universal Classifiers with Natural Language Inference.
\emph{arXiv preprint arXiv:2312.17543}.
\url{https://arxiv.org/abs/2312.17543}.

\bibitem[Mohammadi et al.(2025)]{mohammadi2025}
Seyedali Mohammadi, Bhaskara Hanuma Vedula, Hemank Lamba, Edward Raff, Ponnurangam Kumaraguru, Francis Ferraro, and Manas Gaur. 2025.
Do LLMs Adhere to Label Definitions? Examining Their Receptivity to External Label Definitions.
In \emph{EMNLP}, pages 32380--32393.
\url{https://aclanthology.org/2025.emnlp-main.1648/}.

\bibitem[Pezeshkpour and Hruschka(2024)]{pezeshkpour2024}
Pouya Pezeshkpour and Estevam Hruschka. 2024.
Large Language Models Sensitivity to The Order of Options in Multiple-Choice Questions.
In \emph{Findings of NAACL}, pages 2006--2017.
\url{https://aclanthology.org/2024.findings-naacl.130/}.

\bibitem[Qwen Team(2026)]{qwen2026}
Qwen Team. 2026.
Qwen3.5 model cards.
\emph{Hugging Face model release}.
\url{https://huggingface.co/Qwen/Qwen3.5-4B}.

\bibitem[Reimers and Gurevych(2019)]{reimers2019sbert}
Nils Reimers and Iryna Gurevych. 2019.
Sentence-BERT: Sentence Embeddings using Siamese BERT-Networks.
In \emph{EMNLP-IJCNLP}, pages 3982--3992.
\url{https://aclanthology.org/D19-1410/}.

\bibitem[Schick and Sch{\"u}tze(2021)]{schick2021}
Timo Schick and Hinrich Sch{\"u}tze. 2021.
Exploiting Cloze-Questions for Few-Shot Text Classification and Natural Language Inference.
In \emph{EACL}, pages 255--269.
\url{https://aclanthology.org/2021.eacl-main.20/}.

\bibitem[Stepanov et al.(2025)]{stepanov2025}
Ihor Stepanov, Mykhailo Shtopko, Dmytro Vodianytskyi, Oleksandr Lukashov, Alexander Yavorskyi, and Mykyta Yaroshenko. 2025.
GLiClass: Generalist Lightweight Model for Sequence Classification Tasks.
\emph{arXiv preprint arXiv:2508.07662}.
\url{https://arxiv.org/abs/2508.07662}.

\bibitem[Wang et al.(2024)]{wang2024}
Xinpeng Wang, Chengzhi Hu, Bolei Ma, Paul R{\"o}ttger, and Barbara Plank. 2024.
Look at the Text: Instruction-Tuned Language Models are More Robust Multiple Choice Selectors than You Think.
In \emph{COLM}.
\url{https://arxiv.org/abs/2404.08382}.

\bibitem[Wei et al.(2023)]{wei2023}
Jerry Wei, Le Hou, Andrew Lampinen, Xiangning Chen, Da Huang, Yi Tay, Xinyun Chen, Yifeng Lu, Denny Zhou, Tengyu Ma, and Quoc Le. 2023.
Symbol tuning improves in-context learning in language models.
In \emph{EMNLP}, pages 968--979.
\url{https://aclanthology.org/2023.emnlp-main.61/}.

\bibitem[Xiao et al.(2024)]{xiao2024cpack}
Shitao Xiao, Zheng Liu, Peitian Zhang, Niklas Muennighoff, Defu Lian, and Jian-Yun Nie. 2024.
C-Pack: Packed Resources For General Chinese Embeddings.
In \emph{SIGIR}.
\url{https://doi.org/10.1145/3626772.3657878}.

\bibitem[Xue et al.(2024)]{xue2024}
Mengge Xue, Zhenyu Hu, Liqun Liu, Kuo Liao, Shuang Li, Honglin Han, Meng Zhao, and Chengguo Yin. 2024.
Strengthened Symbol Binding Makes Large Language Models Reliable Multiple-Choice Selectors.
In \emph{ACL}, pages 4331--4344.
\url{https://aclanthology.org/2024.acl-long.237/}.

\bibitem[Yan et al.(2026)]{yan2026}
Lian Yan, Haotian Wang, Chen Tang, Haifeng Liu, Tianyang Sun, Liangliang Liu, Yi Guan, and Jingchi Jiang. 2026.
AgriEval: A Comprehensive Chinese Agricultural Benchmark for Large Language Models.
\emph{Proceedings of the AAAI Conference on Artificial Intelligence}, 40(40):34205--34213.
\url{https://doi.org/10.1609/aaai.v40i40.40716}.

\bibitem[Yang et al.(2022)]{yang2022}
Ping Yang, Junjie Wang, Ruyi Gan, Xinyu Zhu, Lin Zhang, Ziwei Wu, Xinyu Gao, Jiaxing Zhang, and Tetsuya Sakai. 2022.
Zero-Shot Learners for Natural Language Understanding via a Unified Multiple Choice Perspective.
In \emph{EMNLP}, pages 7042--7055.
\url{https://aclanthology.org/2022.emnlp-main.474/}.

\bibitem[Yin et al.(2019)]{yin2019benchmark}
Wenpeng Yin, Jamaal Hay, and Dan Roth. 2019.
Benchmarking Zero-shot Text Classification: Datasets, Evaluation and Entailment Approach.
In \emph{EMNLP-IJCNLP}, pages 3914--3923.
\url{https://aclanthology.org/D19-1404/}.

\bibitem[Yin et al.(2023)]{yin2023}
Fan Yin, Jesse Vig, Philippe Laban, Shafiq Joty, Caiming Xiong, and Chien-Sheng Wu. 2023.
Did You Read the Instructions? Rethinking the Effectiveness of Task Definitions in Instruction Learning.
In \emph{ACL}, pages 3063--3079.
\url{https://aclanthology.org/2023.acl-long.172/}.

\bibitem[Zaratiana et al.(2025)]{zaratiana2025}
Urchade Zaratiana, Gil Pasternak, Oliver Boyd, George Hurn-Maloney, and Ash Lewis. 2025.
GLiNER2: Schema-Driven Multi-Task Learning for Structured Information Extraction.
In \emph{EMNLP: System Demonstrations}, pages 130--140.
\url{https://aclanthology.org/2025.emnlp-demos.10/}.

\bibitem[Zhang et al.(2024)]{zhang2024}
Hang Zhang, Jiawei Sun, Renqi Chen, Wei Liu, Zhonghang Yuan, Xinzhe Zheng, Zhefan Wang, Zhiyuan Yang, Hang Yan, Hansen Zhong, Xiqing Wang, Wanli Ouyang, Fan Yang, and Nanqing Dong. 2024.
Empowering and Assessing the Utility of Large Language Models in Crop Science.
In \emph{NeurIPS, Datasets and Benchmarks Track}.

\bibitem[Zhao et al.(2023)]{zhao2023}
Biao Zhao, Weiqiang Jin, Javier Del Ser, and Guang Yang. 2023.
ChatAgri: Exploring potentials of ChatGPT on cross-linguistic agricultural text classification.
\emph{Neurocomputing}, 557:126708.
\url{https://doi.org/10.1016/j.neucom.2023.126708}.

\bibitem[Zheng et al.(2024)]{zheng2024}
Chujie Zheng, Hao Zhou, Fandong Meng, Jie Zhou, and Minlie Huang. 2024.
Large Language Models Are Not Robust Multiple Choice Selectors.
In \emph{ICLR}.
\url{https://arxiv.org/abs/2309.03882}.
\bibitem[Devane et al.(2025)]{devane2025}
Vijay Devane, Mohd Nauman, Bhargav Patel, Aniket Mahendra Wakchoure, Yogeshkumar Sant, Shyam Pawar, Viraj Thakur, Ananya Godse, Sunil Patra, Neha Maurya, Suraj Racha, Nitish Kamal Singh, Ajay Nagpal, Piyush Sawarkar, Kundeshwar Vijayrao Pundalik, Rohit Saluja, and Ganesh Ramakrishnan. 2025.
BhashaBench V1: A Comprehensive Benchmark for the Quadrant of Indic Domains.
\emph{arXiv preprint arXiv:2510.25409}.
\url{https://arxiv.org/abs/2510.25409}.
\bibitem[Li et al.(2023)]{li2023cmmlu}
Haonan Li, Yixuan Zhang, Fajri Koto, Yifei Yang, Hai Zhao, Yeyun Gong, Nan Duan, and Timothy Baldwin. 2023.
CMMLU: Measuring massive multitask language understanding in Chinese.
\emph{arXiv preprint arXiv:2306.09212}.
\url{https://arxiv.org/abs/2306.09212}.

\bibitem[Huang et al.(2023)]{huang2023ceval}
Yuzhen Huang, Yuzhuo Bai, Zhihao Zhu, Junlei Zhang, Jinghan Zhang, Tangjun Su, Junteng Liu, Chuancheng Lv, Yikai Zhang, Jiayi Lei, Yao Fu, Maosong Sun, and Junxian He. 2023.
C-Eval: A Multi-Level Multi-Discipline Chinese Evaluation Suite for Foundation Models.
\emph{Advances in Neural Information Processing Systems}.
\url{https://arxiv.org/abs/2305.08322}.

\bibitem[Tam et al.(2024)]{tam2024tmmlu}
Zhi-Rui Tam, Ya-Ting Pai, Yen-Wei Lee, Jun-Da Chen, Wei-Min Chu, Sega Cheng, and Hong-Han Shuai. 2024.
An Improved Traditional Chinese Evaluation Suite for Foundation Model.
\emph{arXiv preprint arXiv:2403.01858}.
\url{https://arxiv.org/abs/2403.01858}.

\bibitem[Ying et al.(2025)]{ying2025seedbench}
Jie Ying, Zihong Chen, Zhefan Wang, Wanli Jiang, Chenyang Wang, Zhonghang Yuan, Haoyang Su, Huanjun Kong, Fan Yang, and Nanqing Dong. 2025.
SeedBench: A Multi-task Benchmark for Evaluating Large Language Models in Seed Science.
\emph{Proceedings of the 63rd Annual Meeting of the Association for Computational Linguistics (Volume 1: Long Papers), pages 31395--31449}.
\url{https://aclanthology.org/2025.acl-long.1516/}.

\bibitem[D'Souza(2024)]{dsouza2024agriner}
Jennifer D'Souza. 2024.
Agriculture Named Entity Recognition---Towards FAIR, Reusable Scholarly Contributions in Agriculture.
\emph{Knowledge, 4(1):1--26}.
\url{https://www.mdpi.com/2673-9585/4/1/1}.

\bibitem[Abdelmalak et al.(2025)]{abdelmalak2025fairagro}
Abanoub Abdelmalak, Gabriel Schneider, Heike Riegler, Kristin Meier, Xenia Specka, Nikolai Svoboda, Murtuza Husain, and Juliane Fluck. 2025.
FAIRagro NER Dataset: Increasing FAIRness of FAIRagro Data Through AI-Supported Metadata Enrichment.
\emph{Fachrepositorium Lebenswissenschaften}.
\url{https://doi.org/10.4126/FRL01-6526458}.

\bibitem[Koopman et al.(2022)]{koopman2022agask}
Bevan Koopman, Ahmed Mourad, Hang Li, Anton van der Vegt, Shengyao Zhuang, Simon Gibson, Yash Dang, David Lawrence, and Guido Zuccon. 2022.
AgAsk: An Agent to Help Answer Farmer's Questions From Scientific Documents.
\emph{arXiv preprint arXiv:2212.10762}.
\url{https://arxiv.org/abs/2212.10762}.

\bibitem[MSU-CECO(2025)]{dewlogiqcard}
MSU-CECO. 2025.
DEW-LogiQ dataset card.
\emph{Hugging Face dataset documentation}, repository released 23 October 2025; updated 13 December 2025.
\url{https://huggingface.co/datasets/msu-ceco/dew-logiq}.

\bibitem[Shah(2026)]{agrireasoncard}
Jainam Shah. 2026.
AgriReason-4B: An Open Agronomic Advisory Model.
\emph{Hugging Face model card}, revision \nolinkurl{f255af3513e8dba910025ce38a375482544e6a0d}.
\url{https://huggingface.co/Jainamshahhh/agrireason-4b}.

\bibitem[TypeSafe AI(2026)]{typesafejev}
TypeSafe AI. 2026.
Jev models, versions, and Choice interface.
\emph{Official model documentation}, accessed 11 October 2026.
\url{https://docs.typesafe.ai/models}.
\url{https://docs.typesafe.ai/primitives/choice}.

\bibitem[Houxx(2026)]{agrichncard}
Houxx. 2026.
AgriCHN-2023 dataset and documentation.
\emph{Hugging Face dataset mirror}, released 21 April 2026. Dataset DOI: \url{https://doi.org/10.57967/hf/8798}.
\url{https://huggingface.co/datasets/Houxx/agriCHN}.

\bibitem[AI4Agr(2024)]{crop_resources}
AI4Agr. 2024.
CROP dataset and benchmark resources.
\emph{Official dataset cards}, repositories released 13 June 2024; updated 2 December 2024.
\url{https://huggingface.co/datasets/AI4Agr/CROP-dataset}.
\url{https://huggingface.co/datasets/AI4Agr/CROP-benchmark}.

\bibitem[Almeida(2026)]{almeida2026jev}
Diogo Almeida. 2026.
Introducing System One Models \& Jev.
\emph{TypeSafe AI Blog}.
\url{https://typesafe.ai/blog/introducing-system-one-models-and-jev}.

\bibitem[Cropin AI Lab(2024)]{cropin2024akshara}
Cropin AI Lab. 2024.
Akshara 1.0: Agricultural language model.
\emph{Official model card and release description}.
\url{https://huggingface.co/cropinailab/aksara_v1}.
\url{https://www.cropin.com/blogs/explore-aksara/}.

\end{thebibliography}
\end{document}